\pdfoutput=1

\documentclass[11pt]{article}

\usepackage[final]{acl}

\usepackage{times}
\usepackage{latexsym}

\usepackage[T1]{fontenc}

\usepackage[utf8]{inputenc}

\usepackage{inconsolata}

\usepackage{microtype}
\usepackage{graphicx}
\usepackage{subfigure}
\usepackage{booktabs} 
\usepackage{hyperref}

\usepackage{amsmath}
\usepackage{amssymb}
\usepackage{mathtools}
\usepackage{amsthm}

\usepackage[textsize=small]{todonotes}

\usepackage[capitalize,noabbrev]{cleveref}
\theoremstyle{plain}

\theoremstyle{definition}

\theoremstyle{remark}

\title{ACING: Actor-Critic for Instruction Learning in Black-Box LLMs}
 \author{Salma Kharrat\\
  \And
  Fares Fourati \\ \\
   KAUST \\
  \texttt{\{salma.kharrat, fares.fourati\}@kaust.edu.sa} \\
  \And
  Marco Canini \\}

\begin{document}
\maketitle

\begin{abstract}
The effectiveness of Large Language Models (LLMs) in solving tasks depends significantly on the quality of their instructions, which often require substantial human effort to craft. This underscores the need for automated instruction optimization. However, optimizing instructions is particularly challenging when working with black-box LLMs, where model parameters and gradients are inaccessible. We introduce ACING, an actor-critic reinforcement learning framework that formulates instruction optimization as a stateless, continuous-action problem, enabling exploration of infinite instruction spaces using only black-box feedback. ACING automatically discovers prompts that outperform human-written prompts in 76\% of instruction-induction tasks, with gains of up to 33 points and a 10-point median improvement over the best automatic baseline in 33 tasks spanning instruction-induction, summarization, and chain-of-thought reasoning. Extensive ablations highlight its robustness and efficiency. An implementation of ACING is available at \url{https://github.com/salmakh1/ACING}.
\end{abstract}

\section{Introduction}

Large Language Models (LLMs) have demonstrated impressive capabilities across tasks like classification, summarization, and reasoning \cite{cobbe2021training, touvron2023llama, zhao2023survey}. A key driver of this success is their ability to follow natural language instructions, commonly known as prompts \cite{chen2023you, wei2022chain, liu2023prompt-survey}. Yet crafting effective prompts remains labor-intensive and brittle, particularly in black-box settings where model internals are inaccessible.

A core challenge in prompt engineering lies in the extreme sensitivity of LLMs to subtle linguistic variations. Even minor changes in phrasing can lead to substantial performance shifts. As shown in \cref{tab:prompt_sensitivity}, a minor wording change yields a 12-point gain (using GPT-4o \cite{hurst2024gpt}) despite preserving semantics. Such prompt sensitivity is pervasive, underscoring the need for robust, automated prompt optimization methods.

Recent work has explored automating prompt design to reduce human effort \cite{reynolds2021manual, mishra2021reframing}. Soft prompting techniques \cite{li2021prefix, lester2021power} and heuristic-based discrete search methods \cite{zhou2023large, pryzant2023automatic} have shown promise, yet each faces key limitations in black-box settings. Soft prompts require access to the model internals, limiting them to white-box scenarios. Discrete search methods, in turn, often struggle to explore vast and nuanced instruction spaces efficiently. More recent hybrid methods combine white-box prompt generation with black-box evaluation \cite{chen2023instructzero, instinct}, but typically rely on finite candidate pools or rigid reward assumptions, constraining their applicability.

To overcome these limitations, we introduce \textsc{ACING}, an actor-critic reinforcement learning (RL) framework for automated instruction optimization in black-box LLM settings. \textsc{ACING} formulates prompt optimization as a stateless, continuous-action RL problem within a continuum bandit environment. By learning a latent instruction space through an off-policy actor-critic algorithm, ACING efficiently explores infinite instruction candidates, using only black-box feedback for evaluation. A frozen white-box model is used as a decoder to convert latent vectors into discrete prompts, enabling both scalability and linguistic richness. The generated instructions are not only high-performing but also naturally interpretable, clear, and semantically aligned with the tasks, making them suitable for practical deployment.
To effectively guide exploration under tight query budgets, \textsc{ACING} leverages entropy-regularized policy optimization, encouraging diversity in sampled prompts and improving the likelihood of discovering high-performing instructions.

\begin{table}[t]
\centering
\small
\resizebox{0.45\textwidth}{!}{
\begin{tabular}{clc}
\toprule
\textbf{Source} & \textbf{Instruction} & \textbf{Score} \\
\midrule
Human & Write a word that means the opposite of the input word. &  0.70 \\
ACING & Take a word and change it to its opposite. &  \textbf{0.82} \\
\bottomrule
\end{tabular}}
\caption{\small Our prompt vs human on GPT-4o for antonym task.}
\vspace{-0.4cm}
\label{tab:prompt_sensitivity}
\end{table}

Notably, to our knowledge, ACING is the first approach to apply off-policy, continuous-action actor-critic reinforcement learning to instruction learning in black-box LLMs, using lightweight actor and critic neural networks (rather than LLMs), making the approach both efficient and widely applicable.

We evaluate ACING against state-of-the-art techniques, including Bayesian optimization, contextual bandits, and evolutionary strategies. Empirical results across 33 diverse tasks—including instruction induction, reasoning (including zero-shot Chain-of-Thought (CoT)), semantics, syntax, phonetics, translation, summarization, and code understanding—show that \textsc{ACING} outperforms both human-written prompts and strong automated baselines. Notably, it surpasses human instructions in 76\% of instruction induction tasks, achieving gains of up to 33 points and a median improvement of 10 points over the best automatic baselines.

In summary, our contributions are threefold:

\textbf{(1) An RL formulation of instruction learning:} We propose a continuous-action actor-critic RL approach for instruction optimization in black-box LLMs, enabling scalable exploration of infinite instruction spaces using entropy-regularized policies.

\textbf{(2) Comprehensive validation.} Across 33 diverse tasks, including instruction induction, zero-shot CoT reasoning, and summarization, \textsc{ACING} consistently outperforms both human-written and strong automated instructions, achieving statistically significant gains. 

\textbf{(3) In-depth analysis and insights:} Through extensive ablation studies, we analyze the impact of latent dimensionality, exemplar configuration, decoder architecture, and optimization budget, highlighting the robustness of the \textsc{ACING} approach. Additionally, human evaluation and automated readability analysis confirm the clarity and semantic faithfulness of the generated instructions.


\begin{figure*}[t]
    \centering
\includegraphics[scale=0.34]{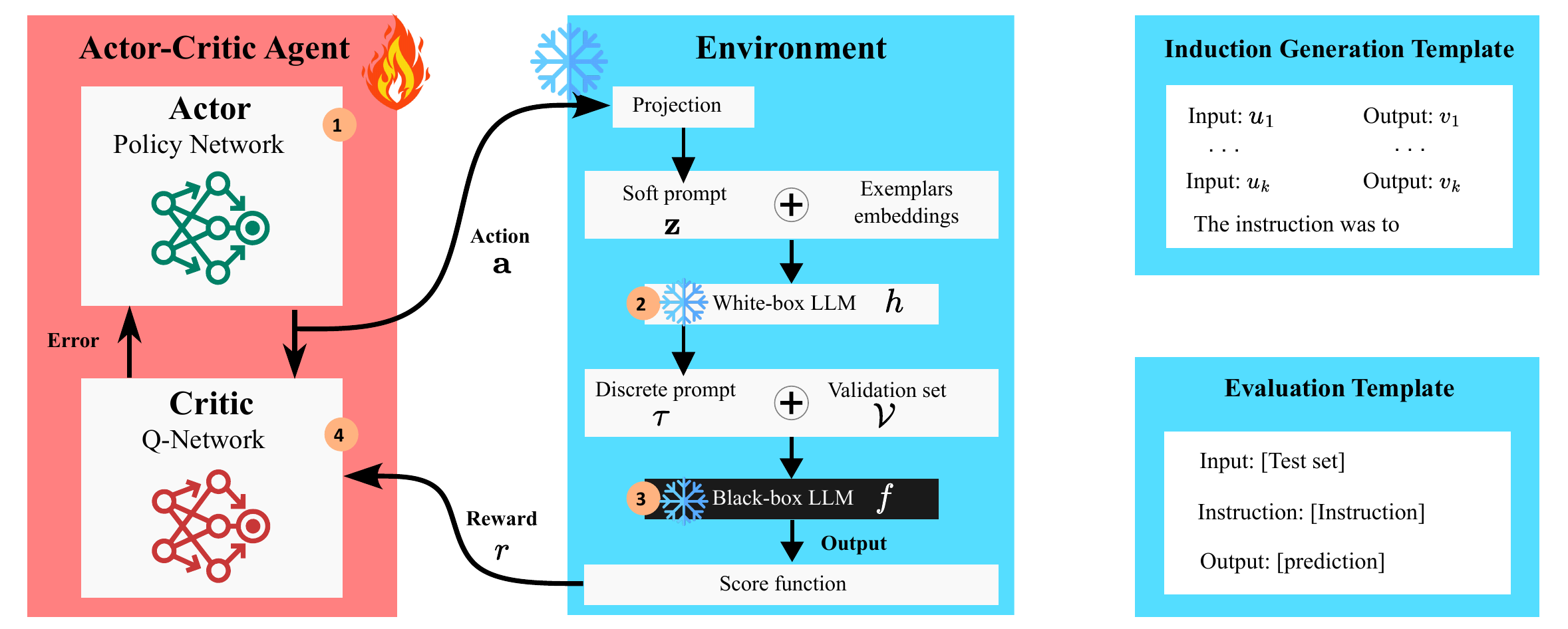}
    \caption{\small \textbf{Pipeline of ACING.}  At each iteration, a soft prompt and task exemplars are fed to the white-box model to generate an instruction. This instruction queries the black-box LLM, whose outputs are scored. The resulting score is returned to the agent as a reward, which is used to update its networks and adjust its policy. Both LLMs remain frozen throughout.}
    
    \label{fig:ACING}
\end{figure*}

\section{Problem Formulation}

\subsection{Problem: Prompt Optimization for Black-Box LLMs}

We aim to improve the performance of a black-box LLM, denoted by $f$, which can only be accessed through its API, while its internal parameters remain unknown. Given a task represented by an (unknown) distribution $(x, y) \sim \mathcal{D}$—where $x$ denotes possible inputs and $y$ the corresponding correct outputs—our goal is to find the optimal prompt $\tau^\star$ that maximizes the likelihood of $f$ producing correct outputs for a given task. This is evaluated using a scoring function $q(\cdot, \cdot) \in [0, 1]$. 

The black-box model $f$ processes an input formed by concatenating ($\oplus$) the prompt $\tau$ with the sentence $x$, producing a predicted output $\hat{y} = f(\tau \oplus x)$. 
More formally, the objective is to maximize the expected score of the LLM in solving the task represented by the distribution $\mathcal{D}$, defined as:
\begin{equation}
\small
\label{discrete_problem}
\max_\tau \quad \mathbb{E}_{(x,y)\sim \mathcal{D}} \left[q(y, f(\tau \oplus x))\right].   
\end{equation}
We utilize a validation dataset $\mathcal{V} = \{(x_j, y_j)\}_{j=1}^{m}$, where each pair consists of an input sentence $x_j$ and its corresponding ground truth output $y_j$. Our objective is to find the prompt that maximizes the scoring function $q(\cdot, \cdot)$ across the validation dataset, where $q(\hat{y}_j, y_j)$ measures the quality of the predicted output $\hat{y}_j$ against the true output $y_j$. Thus, our objective becomes finding the prompt $\tau^\star$ that maximizes the average score over the validation set $\mathcal{V}$. The derived prompt $\tau^\star$ is then evaluated on a separate test set $\mathcal{T} = \{(x^\prime_j, y^\prime_j)\}_{j=1}^{m^\prime}$ to assess its generalization performance.

\subsection{Reformulating Discrete Prompt Search as a Continuous Optimization Problem}

Directly optimizing the prompt $\tau$ for the black-box LLM model $f$ presents substantial challenges due to the discrete combinatorial nature of token selection in $\tau$. To mitigate this challenge, similar to prior approaches \cite{instruct, instinct, hu2024localized} we employ a publicly available, open-source white-box model, represented by $h$, and introduce a \emph{soft prompt} vector $\mathbf{z} \in \mathbb{R}^d$, which is a continuous $d$-dimensional vector representing the token embedding of a set of virtual tokens.
The white-box model $h$, which remains entirely frozen, with no training or gradient updates, serves as a proxy mapping $\mathbf{z}$ into a discrete prompt $\tau$ for the black-box LLM. 

Given a dataset of exemplars, $\mathcal{E} = \{ (u_j, v_j) \}_{j=1}^k$, where each pair $(u_j, v_j)$ defines input-output text sequences that exemplify a downstream task and the vector $\mathbf{z}$, their concatenation is input to the white-box model, generating the discrete prompt 
$\tau(\mathbf{z}) = h(\mathbf{z}, \mathcal{E})$. 
This generated prompt $\tau(\mathbf{z})$ is prepended to a test input $x_j$ from the validation set $\mathcal{V}$, and the combined input is provided to the black-box LLM $f$ to generate an output $\hat{y}_j = f(\tau(\mathbf{z}) \oplus x_j)$. The output $\hat{y}_j$ is then evaluated using the scoring function $q(\hat{y}_j, y_j)$. By using a fixed set of exemplars $\mathcal{E}$, the original discrete problem (Eq. (\ref{discrete_problem})) of finding the optimal prompt $\tau$ is effectively transformed into a continuous optimization problem over the soft prompt vector $\mathbf{z}$, as follows:
\begin{equation}
\small
\label{continious_problem}
\max_{\mathbf{z} \in \mathbb{R}^d} \quad \mathbb{E}_{(x,y)\sim \mathcal{D}} \left[q(y, f(\tau(\mathbf{z}) \oplus x))\right].   
\end{equation}
The soft prompt $\mathbf{z}$ is typically high-dimensional. Therefore, we employ random projection techniques to reduce the input dimension as done in prior works \cite{instruct, instinct}. Specifically, we sample a matrix $P \in \mathbb{R}^{d \times d'}$ with entries from $\text{Uniform}(-1, 1)$, and optimize a lower-dimensional vector $\mathbf{a} \in [0, 1]^{d'}$. The soft prompt is then given by $\mathbf{z} = P\mathbf{a}$, transforming the original problem into optimization over a compact and continuous space as follows:

\begin{equation}
\small
\label{continious_problem_}
\max_{\mathbf{a} \in \mathbb{R}^{d^{'}}} \quad \mathbb{E}_{(x,y)\sim \mathcal{D}} \left[q(y, f(\tau(P\mathbf{a}) \oplus x))\right].   
\end{equation}

This optimization problem is central to our framework. The following section elaborates on our approach to solving it. For clarity, we summarize all symbols used in our actor-critic framework in Table~\ref{tab:notations}.

\begin{figure*}[t]
    \centering
\includegraphics[scale=0.44]{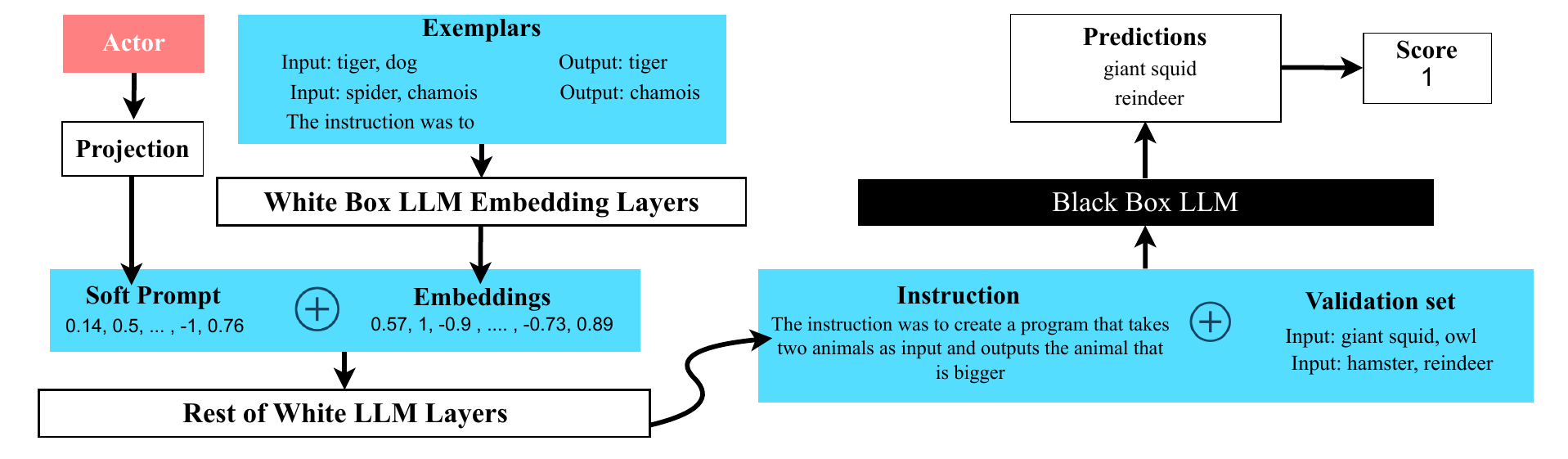}
    \caption{\small Illustration of the prompt generation and testing inside the environment using the \emph{larger\_animal} dataset as an example.}
    \label{fig:animal_ACING}
\end{figure*}

\section{Framework for Instruction Learning}

We formulate the problem of prompt learning for black-box LLMs as a RL problem, where the agent explores an infinite instruction space by sampling continuous actions $\mathbf{a} \in [0,1]^{d^{'}}$. 
Each action corresponds to a latent prompt representation, which is mapped—via a fixed projection matrix and a decoder—into a discrete instruction. This instruction is then evaluated by the black-box LLM on a validation set to produce a reward indicating task performance. The stateless, stochastic setup places the problem in the continuum bandit regime, unlike traditional discrete multi-armed bandits \cite{slivkins2019introduction, lattimore2020bandit}.

\subsection{Actor-Critic for Instruction Optimization}

To efficiently learn in a high-dimensional and limited-feedback setting, we propose a stateless, off-policy actor-critic framework inspired by recent advances in continuous control \cite{konda1999actor, pmlr-v48-mniha16, haarnoja2018soft, haarnoja2018softapp}. Our framework comprises lightweight networks: (1) a policy network (actor) 
$\pi(\cdot; \theta)$, which outputs latent action vectors corresponding to soft prompts, and (2) two critics to compute a value function $Q(\mathbf{a}) = \min\{Q_{\mathbf{w_1}}(\mathbf{a}), Q_{\mathbf{w_2}}(\mathbf{a})\}$, which estimates the expected reward of a given action. Formally, the policy $\pi(\cdot; \theta)$ is modeled as a Gaussian distribution with a diagonal covariance matrix. The actor network outputs the mean and log standard deviation for each action dimension, and actions are sampled using the reparameterization trick.
The critics, $Q_{\mathbf{w_1}}(.)$ and $Q_{\mathbf{w_2}}(.)$, are trained to minimize the mean squared error between predicted and observed rewards:
\begin{equation}
\small
\label{critic_loss}
 \min_{\mathbf{w}} J_Q(\mathbf{w}) \triangleq \mathbb{E}_{\mathbf{a} \sim \mathcal{D}}\left[\frac{1}{2}\left(Q_{\mathbf{w}}\left(\mathbf{a}\right)-r\left(\mathbf{a}\right) \right)^2\right], 
\end{equation}
which can be optimized with stochastic gradients
\begin{equation}
\small
\label{critic_gradient}
\hat{\nabla}_{\mathbf{w}} J_Q(\mathbf{w})=\nabla_{\mathbf{w}} Q_{\mathbf{w}}\left(\mathbf{a}_t\right)\left(Q_{\mathbf{w}}\left(\mathbf{a}_t\right)-r\left(\mathbf{a}_t\right)\right).
\end{equation}

Training alternates between actor and critic updates. Despite the lack of state transitions, this architecture proves effective: the critic generalizes reward signals in prompt space, stabilizes updates by reducing reward variance, and enhances robustness via a twin-critic setup to mitigate overestimation. Empirically, this actor-critic formulation consistently outperforms no-critic baselines.

\subsection{Enhancing Exploration via Entropy Regularization}

In our setting, the agent operates under a fixed evaluation budget of
$T$ queries and must discover the best-performing instruction within this limit. This corresponds to a pure exploration regime, where broad coverage of the action space is essential. While random exploration may seem sufficient, effective learning often requires balancing exploration with exploitation of prior observations to guide the search toward high-reward regions.

To encourage systematic exploration while leveraging past experience, we adopt the maximum entropy reinforcement learning framework \cite{ziebart2010modeling, haarnoja2018soft}. This approach augments the expected reward with a policy entropy term, promoting stochasticity in the actor’s decisions. The objective becomes:
\begin{equation}
\small
\label{policy_loss}
\min_{\theta} J_\pi(\theta) \triangleq \mathbb{E}_{\mathbf{a} \sim \pi(.;\theta)}\left[\alpha \log \left(\pi\left(\mathbf{a}; \theta\right)\right)-Q_{\mathbf{w}}\left(\mathbf{a})\right)\right],
\end{equation}
where $\alpha$ is a temperature coefficient that governs the exploration-exploitation trade-off. A higher $\alpha$ encourages greater policy entropy, while a lower value biases the policy toward exploitation.

This formulation simplifies the soft actor-critic objective \cite{haarnoja2018soft}, omitting state-related components and long-horizon returns, since our environment is stateless and rewards are immediate. The entropy term, $-\mathbb{E}_{\mathbf{a} \sim \pi(\cdot;\theta)}[\log \pi(\mathbf{a})]$, incentivizes the actor to maintain a diverse action distribution, thereby avoiding premature convergence to suboptimal prompts.

To avoid manual tuning of $\alpha$, we follow prior work \cite{haarnoja2018softapp} and adapt it to match a target entropy $H_{\text{target}}$, by minimizing:
\begin{equation}
\small
\label{alpha_loss}
\min_{\alpha} \mathbf{J}_{\alpha} \triangleq -\mathbb{E}_{a \sim \pi(.)} \left[ \alpha \cdot \left( \log \pi(\mathbf{a};\theta) + H_{\text{target}} \right) \right].  
\end{equation}
We use stochastic gradient descent, specifically the Adam optimizer \cite{kingma2014adam, reddi2019convergence}, to jointly update the policy, critic, and entropy temperature.

\subsection{Putting it All Together}
\label{Methodology}

Fig.~\ref{fig:ACING} illustrates our actor-critic framework and its interaction with the environment. Fig.~\ref{fig:animal_ACING} zooms in on an example from the \emph{larger\_animal} dataset.

\textbf{Overview:} In each iteration \( t \leq T\), the actor-critic agent generates a continuous vector ``action'' \(\mathbf{a}\) (step 1). The action is then projected into the appropriate space using a fixed matrix $P$ to obtain \(\mathbf{z}\). The environment then concatenates the projected vector \(\mathbf{z}\) with a set of exemplars' embeddings from \(\mathcal{E}\) and feeds it into a white-box model $h$ (step 2). The white-box model produces a discrete prompt, \(\tau\), which is evaluated using the validation dataset \(\mathcal{V}\) based on the responses from the black-box LLM $f$ (step 3). The black-box LLM's prediction is compared to true labels of the validation examples, and a score function provides a reward, used to update both the critic and actor networks accordingly. 

\textbf{Step \textcircled{{\scriptsize 1}}.} 
The actor outputs the mean and variance of a distribution from which the action, a soft prompt vector $\mathbf{a}$ $\in$ $\mathbb{R}^{d^{'}}$, is sampled. It also computes the log probability, which is key for policy optimization, as shown in Eq.~(\ref{policy_loss}).

\textbf{Step \textcircled{{\scriptsize 2}}.}
As shown in the left side of Fig.~\ref{fig:animal_ACING}, the examples describing the task from the set of exemplars $\mathcal{E}$, along with additional text such as ``The instruction was to,'' are input into the embedding layer of the white-box model to generate continuous vectors (using the instruction generation template in Fig.~\ref{fig:ACING} top right). These continuous vectors are then concatenated with the soft prompt $\mathbf{z}$, projected from the action $\mathbf{a}$. The white-box model layers subsequently process the resulting concatenated vector to produce the discrete prompt $\tau$, suitable for input into the black-box LLM. 

\textbf{Step \textcircled{{\scriptsize 3}}.}
As depicted in the right side of Fig.~\ref{fig:animal_ACING}, for every input $x_i$ in the validation set $\mathcal{V} = \{(x_j, y_j)\}_{j=1}^{m}$, the generated prompt $\tau$ is concatenated to the input sentence $x_i$ and fed to the black-box LLM, which generates an output sentence $\hat{y}_i = f(\tau(\mathbf{z}) \oplus x_i)$.  The output of the black-box LLM is fed into a scoring function $q(\cdot, \cdot)$, which computes the score between the predicted output $\hat{y}_i$ and the true label $y_i$. The overall score is calculated by averaging the scores across all samples, representing the reward: $r = \frac{1}{m} \sum_{i=1}^m q(\hat{y}_i, y_i)$,
where $m$ represents the number of samples.

\textbf{Step \textcircled{{\scriptsize 4}}.}
The critic evaluates the actions taken by the actor using the network $Q_{\textbf{w}}$, which estimates the expected reward for a generated action $\mathbf{a}$ from the policy network $\pi$. Based on the observed reward $r(\mathbf{a})$, the critic updates its network using the loss function (Eq.(\ref{critic_loss})) and gradient (Eq.(\ref{critic_gradient})). The critic’s feedback helps the actor improve its policy by maximizing the reward (Eq.~(\ref{policy_loss})), ensuring a balance between exploration and exploitation.

After \(T\) iterations, the agent returns the best-performing prompt \(\tau^\star\), which is evaluated on the test set \(\mathcal{T}\) in the black-box LLM using the evaluation template shown in Fig.~\ref{fig:ACING} (bottom right).

\begin{table*}[t]
\small
\centering
\resizebox{0.94\textwidth}{!}{
\begin{tabular}{@{}p{0.1\textwidth}@{}p{0.175\textwidth}@{}p{0.37\textwidth}p{0.47\textwidth}p{0.05\textwidth}p{0.06\textwidth}@{}}
\toprule
\textbf{Category}  & \textbf{Task}    &  \textbf{Human Instruction} & \textbf{ACING Instruction (Ours)}  & \textbf{Human} & \textbf{ACING} \\
\midrule
\textit{Spelling}
 & Second\_word\_letter  & Extract the second letter of the input word. &  Input a word and output the letter that corresponds to the second letter in that word
& \textbf{0.96 (0.00)} & 0.92 (0.00) \\

\midrule
\textit{Syntax}                             & Negation     & Negate the input sentence. &  Flip the truth value of the statements in the input & 0.81 (0.00) & \textbf{0.82 (0.00)} \\
\midrule
\textit{Lexical} 

\textit{Semantics} & Antonyms     & Write a word that means the opposite of the input word. & Take a word and change it to its opposite & 0.70 (0.00) & \textbf{0.83 (0.00)}\\
\cmidrule{2-6}
          & Synonyms     & Write a word with a similar meaning to the input word.& Input a word that is a synonym for the word that was output & \textbf{0.14 (0.01)} & 0.13 (0.00)\\
\midrule
\textit{Phonetics}                          & Rhymes       & Write a word that rhymes with the input word. & Input the word that the program thought I was inputting and then output the word that program thought I was inputting & 0.61 (0.01)  & \textbf{1.00 (0.00)}    \\
\midrule
\textit{Semantics} 
& Cause\_and\_effect  & Find which of the two given cause and effect sentences is the cause. & Find the sentence that is the cause of the effect in the pair of sentences &\textbf{0.97 (0.02)} & 0.90 (0.02) \\

\midrule
\textit{Style} & Informal\_to\_formal  & Rephrase the sentence in formal language. & Convert the input into output using the same word order and with the same meaning  & \textbf{0.63 (0.00)} & 0.50 (0.00) \\
\midrule
\textit{Multi-}

\textit{lingual} & Translation\_en-de    & Translate the word into German. & Provide a translation for each word in the English text into German &  0.81 (0.00) & \textbf{0.84 (0.00)}\\
\cmidrule{2-6}
& Translation\_en-es     & Translate the word into Spanish. & Translate the words from English to Spanish, but I noticed that some of the translations are not accurate & \textbf{0.89 (0.00)} & 0.88 (0.00)\\

\cmidrule{2-6}
& Translation\_en-fr   & Translate the word into French. & Create a program that would take an English word as input and output its French equivalent & 0.86 (0.00) & \textbf{0.87 (0.00)}\\
\midrule
\textit{GLUE} & Sentiment 
 & Determine whether a movie review is positive or negative. & Classify each input as positive or negative based on the assessment of the corresponding movie & 0.89 (0.01) & \textbf{0.91 (0.00)} \\
\cmidrule{2-6}
& Sentence\_similarity  & Rate the semantic similarity of two input sentences on a scale of 0 - definitely not to 5 - perfectly. & Find a sentence pair that is probably not similar, and the output is 3 - probably & 0.00 (0.00)& \textbf{0.21 (0.00)} \\
\bottomrule 
 & &  & \textit{median score} & 0.81 & \textbf{0.86 } \\
 & &  & \textit{\# best-performing tasks} & 5 & \textbf{7} \\
 \bottomrule \\
\end{tabular}}
\caption{\small Tasks from the instruction-induction datasets where the human and ACING test scores differed. For each task, we provide the corresponding human instruction as proposed in \cite{honovich-etal-2023-instruction} and our best discovered instruction. We tested these instructions on the test dataset and report the average score (with standard deviation) over 3 repetitions.}
\label{tab:human_performance_main}
\end{table*}

\begin{table*}[h]
    \begin{center}
    \small
    \resizebox{0.83\textwidth}{!}{
    \begin{tabular}{llccccc}
    \hline
    \textbf{Category}          & \textbf{Task}             & \textbf{APE}  & \textbf{EvoPrompt}        & \textbf{InstructZero}   & \textbf{INSTINCT} & \textbf{ACING}\\
    \hline
    \textit{Spelling} 
     & Letters\_list             &   0.59 (0.02)       & 0.97 (0.03)  &    \textbf{1.00 (0.00)}   &    0.99 (0.01)      &       \textbf{1.00 (0.00)}      \\
    & First\_word\_letter       &  0.00 (0.00)  &   \textbf{1.00 (0.00)}    &   \textbf{1.00 (0.00)}                            &    \textbf{1.00 (0.00)}          &  \textbf{1.00 (0.00) }\\
                       & Second\_word\_letter      &  0.00 (0.00)  &  0.63 (0.17)  &    0.35 (0.09)                            &    0.39 (0.28)      &  \textbf{0.70 (0.15) } \\
    \hline
    \textit{Morpho-Syntax} & Negation                  &   0.79 (0.00)  &  \textbf{0.84 (0.02)  } &    0.65 (0.10)   &    0.58 (0.22)     &  0.71 (0.06)    \\
    \hline
    \textit{Lexical Semantics}    
                                & Synonyms                  & 0.14 (0.01)     &  0.19 (0.07)  &    \textbf{0.22 (0.11)}                            &    0.19 (0.08)      &    0.13 (0.02)       \\
                &Word\_unscrambling        &   0.54 (0.00)    &  0.44 (0.06) &    \textbf{0.59 (0.06)}    &    0.54 (0.02)      &     0.50 (0.07)     \\
    \hline
    \textit{Phonetics} & Rhymes                    &  0.59 (0.01)   &  0.52 (0.05)   &    \textbf{0.99 (0.01)}    &    0.36 (0.04)       & 0.57 (0.31)      \\
    \hline
    \textit{Numerical} & Sum                       &   0.87 (0.01)   &  \textbf{1.00 (0.00)}  &   \textbf{1.00 (0.00)}     &    0.70 (0.21)        & \textbf{1.00 (0.00)}    \\
                       & Diff                      &   0.00 (0.00)   &  0.99 (0.01) &   \textbf{1.00 (0.00)}                    &    0.93 (0.09)       &  \textbf{1.00 (0.00)}     \\
    \hline
    \textit{Knowledge} & Larger\_animal            &   0.72 (0.02)   &  0.58 (0.06)  &        0.63 (0.07)                         &    0.81 (0.09)       & \textbf{0.84 (0.07) }      \\
    \hline
    \textit{Cognitive Tasks} & Cause\_and\_effect        &   0.44 (0.09)   &  0.48 (0.10)  &    0.52 (0.09)                                &    0.55 (0.11)                & \textbf{0.69 (0.15)}   \\
                              & Common\_concept           &   0.03 (0.02)   &  0.17 (0.00)   &    0.14 (0.04)                       &    0.09 (0.04)         & \textbf{0.19 (0.05)}    \\
                              & Object\_counting          &    0.30 (0.02)    & \textbf{0.50 (0.06)} &    0.38 (0.06)                             &    0.40 (0.12)      &  0.41 (0.03)\\
                              & Odd\_one\_out             &    0.32 (0.02)  & \textbf{0.64 (0.04)}  &   0.57 (0.02)                              &    0.25 (0.18)        & \textbf{0.64 (0.00)}      \\
                              & Orthography\_starts\_with &    0.23 (0.01)  &  0.47 (0.02)  &    0.41 (0.09)                            & 0.54 (0.06) &      \textbf{0.60 (0.12) }     \\
                              & Taxonomy\_animal          &     0.02 (0.02)  & 0.38 (0.15)  &    0.67 (0.14)                       
                              &    \textbf{0.85 (0.06)}      &    0.71 (0.02) \\
& Auto\_categorization      &   \textbf{0.31 (0.01)}    &  0.20 (0.03) &    0.29 (0.02)                                &    0.07 (0.07)                & 0.29 (0.04)      \\
&Word\_sorting             &    0.58 (0.01)      & 0.01 (0.00)  &  0.64 (0.05)                            &    0.23 (0.20)      &      \textbf{0.70 (0.03)}    \\
    \hline
    \textit{CLUE} & Sentence\_similarity      &   0.00 (0.00)     & 0.05 (0.00) &    0.10 (0.00)                             &    0.00 (0.00)    & \textbf{0.13 (0.07)}  \\
                    
    \hline
    \textit{Translation} 
     &Num\_to\_verbal           &   0.13 (0.02)   & \textbf{1.00 (0.00)}    &   0.99 (0.01)                             &    \textbf{1.00 (0.00)}      &  0.99 (0.01)    \\
                         & Translation\_en-es        & 0.86 (0.01)    &    0.76 (0.00)  &    0.67 (0.24)                             &   \textbf{0.89 (0.00)}   & 0.87 (0.02)   \\
 \hline
    \textit{Style}  & Informal\_to\_formal      &   \textbf{0.57 (0.01)}   &  0.50 (0.02)  &   0.48 (0.02)  &    0.54 (0.09)      & 0.44 (0.05)\\
    \hline
    \textit{Coding} 
                       & Auto\_debugging           &   \textbf{0.25 (0.00)}   &  \textbf{0.25 (0.00)}  &    \textbf{0.25 (0.00)}    &    0.07 (0.07)                & \textbf{0.25 (0.00)}   \\

    \hline
    &median score              &       0.31   & 0.50    &    0.59                                    &     0.54            &        \textbf{0.69}        \\    
    &\# best-performing tasks  &      3     &   7       &       8                                     &       4              &       \textbf{13}        \\
    \hline
    \end{tabular}}
    \end{center}
    \caption{\small Average test performance (with standard deviations) over 3 seeds comparing \textsc{ACING} to APE~\citep{zhou2023large}, EvoPrompt~\citep{guo2023connecting}, InstructZero~\citep{instruct}, and INSTINCT~\citep{instinct} on the 23 most challenging tasks (where at least one method has score < 0.7). Bottom rows show median scores and the number of best-performing tasks.}
    \label{tab:instruction_induction}
\end{table*}

\section{Experiments}
\label{sec:experiments}

We focus on instruction learning for ChatGPT \cite{openai2023chatgpt}, with additional analysis on GPT-4 \cite{openai2023gpt4} and GPT-4o \cite{hurst2024gpt} as representative black-box LLMs.
We conduct instruction induction tasks using 30 datasets spanning several diverse categories from~\citet{honovich-etal-2023-instruction, instruct}, zero-shot CoT reasoning datasets (GSM8K~\cite{cobbe2021training}, AQUARAT~\cite{ling2017program}), and summarization on SAMSum~\cite{gliwa2019samsum}.

In §\ref{main:human_vs_acing}, we compare \textsc{ACING}'s best-learned instructions against human-written prompts from~\cite{honovich-etal-2023-instruction}. We also benchmark, in §\ref{main:acing_vs_others}, against four recent black-box instruction optimization methods: APE~\cite{zhou2023large}, EvoPrompt~\cite{guo2023connecting}, InstructZero~\cite{instruct}, and INSTINCT~\cite{instinct}. Furthermore, we study the interpretability and clarity of generated instructions (§\ref{main:clarity}) and conduct ablation studies (§\ref{ablation_main}) to examine the impact of the different design choices.

To ensure fairness and comparability with other automatic approaches, we adopt a fixed evaluation budget of black-box queries \(T=165\),\footnote{See Appendix~\ref{acing_rewards} for reward plots where \textsc{ACING} often peaks well before the budget is exhausted.} consistent with prior work. Moroever, \textsc{ACING} uses an off-the-shelf, publicly available white-box decoder to map latent vectors to prompts. Moreover, all learning and evaluation occur solely through black-box interactions, which is consistent with other leading methods (e.g., INSTINCT, InstructZero). To ensure fairness, we fix the same decoder Vicuna-13B~\cite{vicuna2023}, as used in their analyses across all methods, while in ablations we compare with WizardLM-13B~\cite{xu2023wizardlm}.

The experimental setup isolates \textsc{ACING}’s core contribution, ensuring that performance gains are not attributable to decoder selection or tuning. Further details, including hyperparameters, are provided in Appendix~\ref{app:details_experimental}.

\subsection{ACING vs. Humans}
\label{main:human_vs_acing}

We compare instructions found by ACING against human-authored prompts from~\citet{honovich-etal-2023-instruction} across a broad range of instruction-induction tasks. Table~\ref{tab:human_performance_main} highlights only those tasks where test performance differed between the two. The full set of tasks is included in Appendix~\ref{sec:best_instruction}.

\textsc{Acing} not only matches but often surpasses human-written instructions—often by substantial margins. For instance, in the Antonyms task, the human instruction (“Write a word that means the opposite of the input word”) scores 0.70. ACING improves this to 0.82 with a more direct and actionable phrasing: “Take a word and change it to its opposite.” The formulation is simpler yet effective.

Consider the rhyming task, where the human instruction—“Write a word that rhymes with the input word”—yields a score of 0.61. ACING significantly improves performance, achieving a perfect score of 1.00 with an alternative phrasing. This highlights ACING’s ability to discover high-performing instructions that align closely with the underlying model behavior.

In the sentence similarity task, where the human instruction results in a score of 0.00, ACING raises performance to 0.21. The highest-scoring ACING instruction introduces a mild incline toward a mid-scale output (“3 - probably”), but its phrasing remains semantically valid and interpretable. A second-best instruction, unbiased, still improves upon the human-written version by 0.14 points.

On average, \textsc{Acing} improves the median task score from 0.81 to 0.86 and outperforms human-written instructions on 7 out of 12 tasks in this subset—underscoring its potential as a practical and effective alternative to manual prompt engineering. 
Further generated instructions are in Appendix~\ref{our_best_instructions_appendix}. Furthermore, extended results with GPT-4o are provided in \cref{tab:gpt4o_results} in \cref{appendix:gpt4o_results}.

\subsection{ACING vs. Other Optimization Methods}
\label{main:acing_vs_others}

\textbf{Instruction-induction datasets:}
In Table \ref{tab:instruction_induction}, we show tasks from \cite{honovich-etal-2023-instruction} where at least one method failed to achieve 70\% score (full results on all 30 tasks can be found in \cref{App:ACING_vs_Other_Methods}). The table shows the average test accuracy (along with the standard deviation) over three independent runs, using three different seeds. For each seed, we selected the best instruction achieved by each method and evaluated it on the testing dataset. The results demonstrate that our method, ACING, outperforms the others, achieving the highest accuracy in 13 out of the remaining 23 tasks, compared to INSTINCT, InstructZERO, EvoPrompt, and APE, which succeeded in 8 tasks or fewer. Additionally, ACING achieves the highest median accuracy across tasks, with a value of 0.69, which is approximately 10 percentage points higher than the best baseline. The score types can be found in Table~\ref{tab:score-types}. Moreover, the best prompt achieved for each task and the corresponding test scores, can be found in Table~\ref{tbl:best_prompt_instruction_induction} in \cref{App:ACING_vs_Other_Methods}. Further analyses on GPT-4 can be found in the Appendix.~\ref{secApp:gpt-4}.

\textbf{CoT datasets.} We evaluate our method on two zero-shot CoT reasoning datasets: GSM8K~\cite{cobbe2021training} and AQUA-RAT~\cite{ling2017program}. Prior work~\cite{kojima2022large} shows that simple CoT prompts can enhance LLM performance. Using 100 steps (and other settings as above), ACING achieves the best results on both, demonstrating strong CoT reasoning capability.

\textbf{Summarization dataset.} We compare the performance of ACING with other methods on summarization tasks using the SAMSum dataset \cite{gliwa2019samsum}. The results, presented in Table \ref{tab:summarization}, show that ACING outperforms the other methods across the three metrics considered: ROUGE-1, ROUGE-2, and ROUGE-L \cite{lin2004rouge}.

\textbf{Statistical significance test.} We conduct a Wilcoxon signed-rank test \cite{wilcoxon1992individual}, a non-parametric test. The results confirm that \textsc{ACING} significantly outperforms all baselines across tasks, with p = 0.0005 (APE), 0.0041 (INSTINCT), 0.0092 (EvoPrompt), and 0.0335 (InstructZero), all below the standard 0.05 threshold, indicating that the observed gains are statistically significant.

\begin{table}[t]
\begin{center}
\resizebox{0.47\textwidth}{!}{
\begin{tabular}{lccccc}
\hline
\textbf{Metric} & \textbf{APE} & \textbf{EvoPompt} & \textbf{InstructZero} & \textbf{INSTINCT} & \textbf{ACING} \\
\hline
ROUGE-1 & 0.35 (0.01) & 0.35 (0.01) & 0.33 (0.00) & 0.36 (0.01) & \textbf{0.37 (0.01)} \\
ROUGE-2 & 0.12 (0.00) & 0.12 (0.00) & 0.11 (0.00) & \textbf{0.14 (0.00)} & \textbf{0.14 (0.00)} \\
ROUGE-L & 0.25 (0.00) & 0.26 (0.00) & 0.24 (0.01) & 0.27 (0.01) & \textbf{0.28 (0.01)} \\
\hline
\end{tabular}
}
\end{center}
\caption{\small Average test performance (and standard deviations) for summarization task using SAMSum dataset.}
\vspace{-0.6cm}
\label{tab:summarization}
\end{table}

\begin{figure}[t]
    \centering
\includegraphics[scale=0.13]{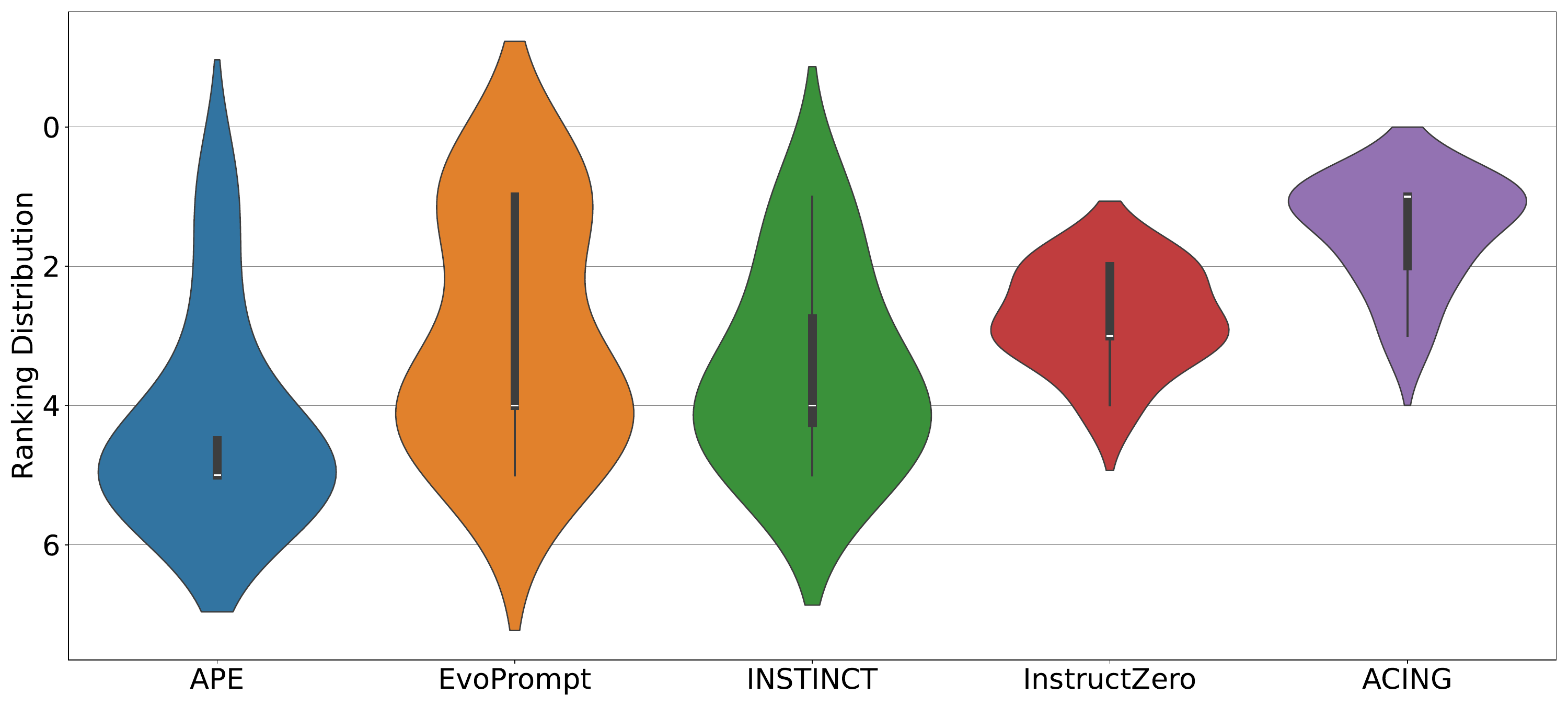}
    \caption{\small Ranking distributions across cognitive tasks for all algorithms, ordered increasing by median rank.}
    \label{fig:violin}
    \vspace{-0.4cm}
\end{figure}

\textbf{Results breakdown.} To better understand ACING's strengths, we grouped the tasks into various categories. On cognitive tasks, requiring complex reasoning, \textsc{ACING} demonstrates the clearest advantage. It ranks first on 5 out of 8 such tasks from Table~\ref{tab:instruction_induction}, including Cause\_and\_effect (0.69, +14\%), Word\_sorting (0.70, +6\%), Odd\_one\_out (0.64, tied), and Orthography\_starts\_with (0.60, +6\%). As shown in Fig.~\ref{fig:violin} (ordered left to right by median ranking), \textsc{ACING} achieves the best median rank and shows a strong skew toward top rankings in this category. Furthermore, ACING shows the best performance in the zero-shot CoT reasoning tasks (\cref{tab:zero-shot-instructions}). In symbolic manipulation, ACING again performs best, achieving perfect scores on Letters\_list and First\_word\_letter, and leading on Second\_word\_letter (0.70 vs. 0.63). This highlights its precision in low-level, structured tasks. ACING is competitive but not dominant in tasks involving lexical semantics, world knowledge, and translation—e.g., Informal\_to\_formal (0.44 vs. 0.57), and Translation\_en-es (0.87 vs. 0.89). Finally, it consistently outperforms the other methods in summarization.

\begin{table*}[t]
\centering
\small
\resizebox{0.85\textwidth}{!}{
\begin{tabular}{l|l|p{8cm}|c}
\hline
\textbf{Method} & \textbf{Dataset} & \textbf{Best Zero-Shot Instruction} & \textbf{Score} \\
\hline
\cite{kojima2022large} & GSM8K & Let's think step by step. & 0.72 \\
\textbf{INSTINCT} \cite{instinct} & GSM8K & Let's use our creativity to find the solution. & 0.75 \\
\textbf{ACING} (Ours) & GSM8K & Let's use our math skills to conquer this challenge. & \textbf{0.76} \\
\hline
\hline
\cite{kojima2022large} & AQUA-RAT & Let's think step by step. & 0.59 \\
\textbf{INSTINCT} \cite{instinct} & AQUA-RAT & Let's break it down. & 0.59 \\
\textbf{ACING} (Ours) & AQUA-RAT & Let's use the power of substitution to solve this problem. & \textbf{0.63} \\
\hline
\end{tabular}
}
\caption{Best zero-shot instructions and corresponding scores for different methods and datasets.
}
\label{tab:zero-shot-instructions}
\end{table*}

\subsection{Instruction Clarity and Readability}
\label{main:clarity}

To evaluate the interpretability and clarity of generated instructions, we assess whether \textsc{ACING}-generated prompts are understandable and task-aligned. We conduct a human evaluation with 26 participants who rated the clarity and alignment of generated instructions from \cref{tab:human_performance_main} on a 5-point Likert scale. The results show that participants found the prompts generally clear, with a median score of 3.9/5 (see Appendix~\ref{sec:instruction-clarity} for the protocol and statistics). In parallel, we use standard readability metrics to automatically assess the generated prompts. The results show a median Flesch Reading Ease (FRE) of 70.8, Flesch-Kincaid Grade Level (FKG) of 7.0, and Coleman-Liau Index (CLI) of 7.3. These scores correspond to mid-grade readability (7th–8th grade), indicating that the generated instructions are accessible to a broad user base.

\subsection{Ablation Studies}
\label{ablation_main}
We present detailed ablation studies on key design choices, summarize some below, and provide full results and plots in Appendices~\ref{app:ablation_design}--\ref{different_white_llms}.

\textbf{Use of critics.} We compare our method (with two critics) to variants with a single critic and to a baseline without a critic (policy-gradient). As shown in Appendix~\ref{app:ablation_design}, the two-critic architecture yields the highest accuracy, best stability, and top performance in difficult settings (e.g., cognitive), confirming the value of conservative estimation.

\textbf{Budget efficiency.} While ACING uses a fixed 165-query budget for fair comparison with prior work, it often converges well before the budget is exhausted. As illustrated in Appendix~\ref{acing_rewards}, many tasks reach optimal rewards within 60–80 queries, and some within 10–20, showing strong sample efficiency under constrained settings.

\textbf{White-box model.} Using  WizardLM-13B~\cite{xu2023wizardlm} instead of Vicuna improves median test accuracy by 8 points and increases the number of best-performing tasks (Appendix~\ref{different_white_llms}). This demonstrates that ACING can benefit from stronger decoding models, although it remains effective across architectures.

\textbf{Action dimensionality.} We test latent action sizes $d' \in \{5, 10, 20, 40, 100\}$. Results in Appendix~\ref{different_dimensions} show that $d'=10$ and $d'=20$ perform consistently well, while larger dimensions like $d'=40$ yield improvements on specific tasks. 

\textbf{Number of exemplars.} Using a single exemplar performs surprisingly well and matches the 5-exemplar setup on several tasks (e.g., phonetics and summation). Still, five exemplars offer more consistent gains on complex tasks (\cref{different_exemplars}).

\textbf{Budget splitting.} We explore dividing the query budget into an exploration phase and a final re-ranking phase, where top prompts are re-evaluated multiple times. This two-phase strategy improves median test scores by 5 points and boosts the number of best-performing tasks (\cref{split_budget_appendix}).

\section{Related Work}

Early approaches such as AutoPrompt~\cite{shin2020autoprompt}, FluentPrompt~\cite{shi2022toward}, and soft prompt tuning~\cite{lester2021power, li2021prefix, zhong2021factual} rely on access to model gradients or embeddings, limiting their applicability in black-box settings. Grey-box methods such as BBT and BBTv2~\cite{sun2022black, sun2022bbtv2} and CLIP-tuning~\cite{chai2022clip} relax these constraints by leveraging token embeddings or logits, but are still incompatible with API-only black-box models.

Several recent works frame instruction search as a discrete optimization problem. RLPrompt~\cite{deng2022rlprompt} and Tempera~\cite{zhangtempera} use RL to identify prompts, but assume access to token-level outputs or confidence scores. Alternatively, sampling- and evolution-based strategies such as APE~\cite{zhou2023large}, PromptBreeder~\cite{fernando2024promptbreeder}, EvoPrompt~\cite{guo2023connecting}, PromptWizard~\cite{agarwal2024promptwizard}, and Auto Evol-Instruct~\cite{zeng2024automatic} iteratively generate and refine candidate prompts via LLM sampling or mutation. While effective in some cases, these methods typically rely on large candidate pools or expensive query budgets. 

Zeroth-order optimization (e.g., AIO~\cite{qiautomatic}, ZOPO~\cite{hu2024localized}) estimates gradients without backpropagation, yet incurs high computational cost and query complexity. InstructZero~\cite{instruct} and INSTINCT~\cite{instinct} apply Bayesian Optimization (BO) and NeuralUCB~\cite{zhou2020neural} within a finite action space. In contrast, StablePrompt~\cite{kwon2024stableprompt} uses PPO to stabilize discrete prompt learning, but fine-tune white-box LLMs. Our approach differs by formulating prompt optimization as a continuous-action RL problem, enabling more efficient exploration without gradient access or large models.

Complementary strategies include exemplar selection~\cite{wu2024prompt}, preference-based feedback~\cite{lin2024prompt}, and best-arm identification~\cite{shi2024best}. However, these typically rely on fixed prompt pools or require human preference labels. ACING jointly addresses both the generation and selection in a unified framework, yielding high-performing prompts without pool constraints or external supervision.

\section{Conclusion}

We present \textsc{ACING}, an actor-critic RL framework for prompt optimization in black-box LLMs. By formulating the task as a continuous-action problem, \textsc{ACING} enables exploration of an infinite space through an entropy-regularized policy under strict query constraints. It outperforms strong baselines and surpasses human-written instructions, without any per-task tuning, demonstrating effectiveness, efficiency, and practicality.

\section{Limitations}

While \textsc{ACING} achieves strong and consistent performance across diverse tasks, we do not claim universal superiority on every individual task. Given the stochastic and non-convex nature of prompt optimization—particularly in black-box LLMs under tight query budgets—strong prompts can occasionally arise even from random search or other optimization approaches. However, \textsc{ACING} substantially improves the chances of discovering such prompts through a principled exploration strategy grounded in actor-critic learning.

Similar to previous prior works \citep{instinct, instruct, hu2024localized}, our reliance on a white-box model introduces variability, as model selection affects performance (Appendix~\ref{different_white_llms}). Addressing this limitation would require operating directly on black-box LLMs, which poses challenges due to the large discrete action space. Adapting RL methods for such spaces, as explored in \citet{pmlr-v235-fourati24a}, and proposing hybrid approaches combining soft-prompt and discrete optimization, represent promising directions for future research.

Despite the strong performance achieved with a fixed hyperparameter configuration across all tasks, we acknowledge that domain-specific tuning might yield marginal gains, particularly in edge cases. That said, our goal was to emphasize generality and simplicity without introducing per-task overhead—an important factor for real-world usability.

\section{Ethical Considerations}

\textsc{ACING} automates prompt optimization for LLMs, which can reduce manual effort but may also amplify risks associated with bias, misuse, or harmful content generation. We emphasize that \textsc{ACING} is a general optimization framework and does not guarantee ethical outputs from the underlying LLMs. It should be paired with appropriate content filters and safety mechanisms when deployed in sensitive domains. Optimized prompts could reinforce or amplify existing biases present in the base LLMs. Future work should incorporate fairness-aware reward functions or post-hoc bias mitigation. 

Lastly, our experiments assume legal and ethical API usage, and we caution against applying instruction optimization to restricted or proprietary models without adherence to usage policies and terms of service.

\bibliography{main}

\appendix

\newpage
\section{Table of Notations}
For clarity, we summarize all symbols used in our actor-critic framework in Table~\ref{tab:notations}.

\begin{table*}[h]
\centering
\begin{tabular}{ll}
\hline
\textbf{Symbol} & \textbf{Meaning} \\
\hline
$f$ & Black-box LLM (only accessible via API) \\
$h$ & White-box LLM (frozen, used as decoder) \\
$D$ & Task distribution over input–output pairs $(x,y)$ \\
$V$ & Validation set $\{(x_j, y_j)\}_{j=1}^m$ \\
$T$ & Test set $\{(x'_j, y'_j)\}_{j=1}^{m'}$ \\
$q(\cdot,\cdot)$ & Scoring function measuring output quality \\
$\tau$ & Discrete prompt \\
$\tau(z)$ & Prompt generated from soft prompt $z$ \\
$\tau(Pa)$ & Prompt from projected action vector $a$ \\
$\tau^\star$ & Optimal prompt \\
$z \in \mathbb{R}^d$ & Soft prompt vector (continuous) \\
$a \in [0,1]^{d'}$ & Action (latent prompt representation) \\
$P \in \mathbb{R}^{d \times d'}$ & Random projection matrix \\
$\pi(\cdot;\theta)$ & Policy network (actor) with parameters $\theta$ \\
$Q_{w_1}(a), Q_{w_2}(a)$ & Twin critic networks with parameters $w_1, w_2$ \\
$Q(a)$ & Value estimate: $Q(a) = \min\{Q_{w_1}(a), Q_{w_2}(a)\}$ \\
$J_Q(w)$ & Critic loss: $\mathbb{E}_{a \sim D} \left[\tfrac{1}{2}(Q_w(a)-r(a))^2\right]$ \\
$J_\pi(\theta)$ & Actor loss (entropy-regularized policy objective) \\
$r(a)$ & Reward obtained from the environment \\
$\alpha$ & Temperature coefficient for entropy regularization \\
$H_{\text{target}}$ & Target entropy for adaptive $\alpha$ \\
\hline
\end{tabular}
\caption{Summary of notations used in the ACING actor-critic framework.}
\label{tab:notations}
\end{table*}

\section{Experimental Details}
\label{app:details_experimental}

\subsection{Hyperparameters}
\label{App:hyperparameters}
 Across the diverse tasks, in the main paper, the same hyperparameters were used, which shows that the algorithm generalizes well across the 30 tasks without specifically tuning hyperparameters in each task. A summary of the key parameters can be found in the following Table. 

\begin{table*}[h]
\centering
\begin{tabular}{|c|c|}
\hline
\textbf{Hyperparameter} & \textbf{Choice} \\
\hline
White-box $h$ & Vicuna-13B and WizardLM \\
Actor-network & $(1-1024-256-10)$ \\
Critic-network & $(10-128-128-1)$ \\
Budget $T$ & 165 \\
Intrinsic (action) dimension $d'$ & 10 \\
Number of soft tokens $N_z$ & 5 \\
Soft prompt dimension $d$ & 5120 * $N_z$ \\
Number of exemplars $|\mathcal{E}|$ & 5 \\
Number of tokens generated by Wb & 64 \\
\hline
\end{tabular}
\caption{Key hyperparameters and their values.}
\end{table*}

Furthermore, like previous works, we use their default (tuned) hyperparameters for the results in the main paper, including the intrinsic dimension $d' = 10$ and the number of soft tokens $N_z = 5$. For fairness, we refrain from fine-tuning these parameters for our method and use the same values as in prior works. This ensures that our ACING algorithm searches in the same space, $[0,1]^{10}$, and uses the same total number of queries to the black-box LLM as APE, InstructZero, and INSTINCT for a fair comparison. For each algorithm, after identifying the best instruction using the validation set $\mathcal{V}$, we evaluate the discovered instruction on a separate test set $\mathcal{T}$ and report the test score. 

\subsection{Actor-Critic Details}
Across all the tasks, we used three fully-connected layers for both the actor $(1-1024-256-10)$ and the critics $(10-128-128-1)$ networks, with learning rates fixed at $3 \cdot 10^{-4}$ for each.  We learn the entropy parameter $\alpha$ using a learning rate of $9 \cdot 10^{-4}$. We adopt two independently trained critics and take the minimum of their outputs to mitigate overestimation bias. This conservative approach helps regularize training and improves robustness, particularly in tasks with noisy or sparse rewards. We further validate this choice in our ablation studies with one critic, two critics, and without critics.

\subsection{Metrics}

Table~\ref{tab:score-types} outlines the evaluation metrics used across various task types. Depending on the nature of the task, we adopt different scoring schemes to ensure fair and meaningful evaluation. For most classification and generation tasks, we employ exact match (EM) scoring, which requires the prediction to match the ground truth exactly, making it a stringent yet interpretable metric.

\subsection{Licenses and Terms of Use for Artifacts}

\paragraph{Black-box APIs.} We use the OpenAI API to access ChatGPT \cite{citechatgpt}, GPT-4 \cite{openai2023gpt4}, and GPT-4o \cite{hurst2024gpt} for black-box evaluation. These models are proprietary and accessed via paid API under OpenAI’s terms of service. We do not redistribute or modify these models.

\paragraph{White-box Models.} Our framework uses Vicuna-13B \cite{vicuna2023} (and WizardLM \cite{xu2023wizardlm} for ablations) as a frozen decoder to generate discrete prompts. Both models are publicly available for research use under a non-commercial license. We follow all usage restrictions and do not modify or redistribute these models.

\paragraph{Datasets.} We use publicly available datasets including:
\begin{itemize}
  \item Instruction induction datasets from \citet{honovich-etal-2023-instruction}.
  \item SAMSum \citep{gliwa2019samsum} for summarization.
  \item GSM8K \citep{cobbe2021training} and AQUA-RAT \citep{ling2017program} for reasoning.
\end{itemize}
All datasets are used under their original licenses and are cited appropriately.

\paragraph{Our Code and Models.} We will publicly release our implementation, including the actor-critic training framework and prompt optimization pipeline, under an open-source license (MIT license) upon publication.

\subsection{Data Usage, Privacy, and Safety Considerations}

All datasets for the studied tasks are publicly released and widely used in NLP research. We used them under their respective licenses for academic use, and we cite the original sources in all cases.

\paragraph{Privacy and Anonymization.} None of the datasets we used contain personally identifiable information (PII) to the best of our knowledge. The datasets are either synthetic, anonymized, or derived from public sources with appropriate preprocessing. We did not augment or modify these datasets in ways that would introduce identifying information.

\paragraph{Human Evaluation and Safety.} For our human evaluation of ACING-generated prompts, we used only automatically generated instructions and task templates. Prior to annotation, we reviewed the generated prompts and outputs to ensure that they did not contain any offensive or inappropriate content. The tasks involved abstract or anonymized content (e.g., generic input strings or public task descriptions), and no human names, images, or personal data were included.

\paragraph{Ethical Use.} We ensured that all experiments involving human subjects were conducted ethically and responsibly. Participation in the annotation study was fully voluntary. Annotators were informed about the nature and purpose of the task and could withdraw at any time. The study was limited to rating task-level clarity and alignment, involved no sensitive content, and did not collect any personal or identifiable information. All responses were anonymized, and only aggregate statistics were analyzed. As the study posed no foreseeable risk to participants and did not involve identifiable data, it did not require formal ethics review under our institution’s guidelines.

\subsection{Artifact Documentation}

\paragraph{Task Coverage.} We evaluate \textsc{ACING} on a diverse set of NLP tasks, categorized as follows:
\begin{itemize}
    \item \textbf{Instruction Induction:} A wide range of classification and transformation tasks drawn from \citet{honovich-etal-2023-instruction}, covering syntax, semantics, phonetics, word manipulation, and reasoning.
    \item \textbf{Reasoning:} Zero-shot chain-of-thought reasoning tasks using GSM8K \citep{cobbe2021training} and AQUA-RAT \citep{ling2017program}, which involve multi-step symbolic and arithmetic problem solving.
    \item \textbf{Summarization:} Dialogue-based abstractive summarization using the SAMSum dataset \citep{gliwa2019samsum}.
    \item \textbf{Translation:} Lexical-level English-to-German, English-to-Spanish, and English-to-French tasks to assess multilingual instruction learning.
\end{itemize}

\paragraph{Language Coverage.} Our experiments involve four languages:
\textit{English} (the primary language for most tasks), and \textit{German}, \textit{Spanish}, and \textit{French} in translation tasks.

\paragraph{Linguistic Phenomena.} The tasks cover a broad range of linguistic phenomena, including:
\begin{itemize}
    \item \textbf{Morphosyntax:} e.g., negation, word reordering.
    \item \textbf{Lexical Semantics:} e.g., antonyms, synonyms, word similarity.
    \item \textbf{Discourse and Pragmatics:} e.g., summarization, cause-effect reasoning.
    \item \textbf{Symbolic and Numerical Reasoning:} e.g., math problem solving, program induction.
\end{itemize}

\paragraph{Demographics and Bias.} All datasets used are publicly available academic benchmarks. None were collected or annotated by the authors. To the best of our knowledge, these datasets do not contain personally identifiable information (PII) or explicit demographic attributes. No demographic inference, fairness evaluation, or bias analysis was performed, as our work focuses on instruction optimization and not on social or identity-based NLP tasks.

\paragraph{Human Annotation.} We conducted a human evaluation study to assess the clarity and semantic alignment of instructions generated by \textsc{ACING}. Annotators were presented with anonymized task prompts and model-generated instructions, and asked to rate clarity and alignment on a 5-point Likert scale. The annotation involved 26 in-house participants, all of whom were informed of the task goals and free to opt out at any time. No demographic data was collected. All instructions shown were manually reviewed to ensure that they contained no PII or offensive content. Collected responses were fully anonymized, and no free-form sensitive text was stored. Results are summarized in Section~\ref{main:clarity} and \cref{sec:instruction-clarity}. The full set of instructions used in the study is from \cref{tab:human_performance_main}.

\paragraph{Artifact Availability.} We will release our full codebase under an open-source license (MIT license) upon publication. This includes the reinforcement learning framework, prompt decoding pipeline, and evaluation scripts. All third-party datasets and models used are cited and publicly available under their respective research licenses.

\subsection{Implementation and Compute Details}

\paragraph{Model Sizes.}
Our actor and critic networks are lightweight MLPs with three layers each. The white-box decoder is Vicuna-13B, and the black-box models queried include GPT-3.5 (ChatGPT), GPT-4, and GPT-4o, whose sizes are not publicly disclosed.

\paragraph{Compute Infrastructure.}
We use an internal SLURM cluster for running our experiments. The experiments were done on an ASUS ESC N4A-E11 server. The node has 4 A100 GPUs, an AMD EPYC 7003 series 64 core @ 3.5GHz CPU and 512GB of RAM.
We used one A100, with 2 cores, and required at most 50GB of memory for the experiments.
All black-box LLM queries were performed via API.

\paragraph{Compute Budget.}
Each ACING run used a fixed query budget of 165 black-box API calls. Across 33 tasks and 3 random seeds, this corresponds to a total of approximately 16,000 black-box LLM queries. Training time per task was approximately 10–20 minutes on a single GPU.

\begin{table*}[h]
\centering
\resizebox{\textwidth}{!}{
\begin{tabular}{|l|l|}  
\hline
\textbf{Task} & \textbf{Score Type} \\
\hline
Common\_concept, Informal\_to\_formal & F1 score = $(2 \times \text{precision} \times \text{recall}) / (\text{precision} + \text{recall})$ \\
\hline
Orthography\_starts\_with, Taxonomy\_animal & Set match: prediction set must match ground truth set \\
\hline
Synonyms & In-list match: prediction is correct if it falls within the list of ground truth words \\
\hline
All other datasets (e.g., Antonyms, Translation, Cause\_and\_effect, Diff) & EM score: 1 if exact match, 0 otherwise (letter-by-letter for words) \\
\hline
\end{tabular}}
\caption{Scoring metrics for different tasks.}
\label{tab:score-types}
\end{table*}

\section{ACING vs. Other Optimization Methods}
\label{App:ACING_vs_Other_Methods}

We compare our method against recent baselines on the 30 instruction-induction datasets. The results in Table \ref{tab:app_instruction_induction} show the average test accuracy (along with the standard deviation) over three independent runs, using three different seeds. For each seed, we selected the best instruction achieved by each method and evaluated it on the testing dataset. The table demonstrates that our method, ACING, outperforms others by achieving the highest accuracy in 14 out of 30 tasks, compared to INSTINCT, InstructZERO, EvoPrompt, and APE which succeeded in 8 tasks or less each. Additionally, ACING achieves the highest median accuracy across tasks, with a value of 0.71, which is ~22 percentage points higher than APE. Table~\ref{tbl:best_prompt_instruction_induction} shows the best prompt achieved for each task and corresponding test scores.

\begin{table*}[h]
    \begin{center}
    \resizebox{0.528\textheight}{!}{
    \resizebox{0.99\textwidth}{!}{
    \begin{tabular}{llccccc}
    \hline
    \textbf{Category}          & \textbf{Task}             & \textbf{APE}  & \textbf{EvoPrompt}        & \textbf{InstructZero}   & \textbf{INSTINCT} & \textbf{ACING}\\
    \hline
    \textit{Spelling} 
     & Letters\_list             &   0.59 (0.02)       & 0.97 (0.03)  &    \textbf{1.00 (0.00)}   &    0.99 (0.01)      &       \textbf{1.00 (0.00)}      \\
    & First\_word\_letter       &  0.00 (0.00)  &   \textbf{1.00 (0.00)}    &   \textbf{1.00 (0.00)}                            &    \textbf{1.00 (0.00)}          &  \textbf{1.00 (0.00) }\\
                       & Second\_word\_letter      &  0.00 (0.00)  &  0.63 (0.17)  &    0.35 (0.09)                            &    0.39 (0.28)      &  \textbf{0.70 (0.15) } \\
    \hline
    \textit{Morpho-Syntax} & Singular\_to\_plural      &  \textbf{1.00 (0.00)}  &     \textbf{1.00 (0.00)}   &       0.99 (0.01)                          &    0.95 (0.03)          &  0.95 (0.03)  \\
                            & Active\_to\_passive       &   \textbf{1.00 (0.00)} & 0.99 (0.00) &   0.98 (0.01)                       &     \textbf{1.00 (0.00)}       & \textbf{1.00 (0.00)}     \\ 
                            & Negation                  &   0.79 (0.00)  &  \textbf{0.84 (0.02)  } &    0.65 (0.10)                            &    0.58 (0.22)     &  0.71 (0.06)    \\
    \hline
    \textit{Lexical Semantics} & Antonyms                  &   0.79 (0.02)   &  0.70 (0.01)  &    0.76 (0.00)                                &   \textbf{0.84 (0.01)}       & 0.74 (0.01)                     \\
                                & Synonyms                  & 0.14 (0.01)     &  0.19 (0.07)  &    \textbf{0.22 (0.11)}                            &    0.19 (0.08)      &    0.13 (0.02)       \\
                &Word\_unscrambling        &   0.54 (0.00)    &  0.44 (0.06) &    \textbf{0.59 (0.06)}    &    0.54 (0.02)      &     0.50 (0.07)     \\
    \hline
    \textit{Phonetics} & Rhymes                    &  0.59 (0.01)   &  0.52 (0.05)   &    \textbf{0.99 (0.01)}    &    0.36 (0.04)       & 0.57 (0.31)      \\
    \hline
    \textit{Numerical} & Sum                       &   0.87 (0.01)   &  \textbf{1.00 (0.00)}  &   \textbf{1.00 (0.00)}     &    0.70 (0.21)        & \textbf{1.00 (0.00) }   \\
                       & Diff                      &   0.00 (0.00)   &  0.99 (0.01) &   \textbf{1.00 (0.00)}                    &    0.93 (0.09)       &  \textbf{1.00 (0.00)}     \\
    \hline
    \textit{Knowledge} & Larger\_animal            &   0.72 (0.02)   &  0.58 (0.06)  &        0.63 (0.07)                         &    0.81 (0.09)       & \textbf{0.84 (0.07) }      \\
                       & Periodic\_elements        &  0.99 (0.01)  &   0.92 (0.00)    &           0.96 (0.03)                      &   \textbf{1.00 (0.00)}          & 0.98 (0.00)        \\
    \hline
    \textit{Cognitive Tasks} & Cause\_and\_effect        &   0.44 (0.09)   &  0.48 (0.10)  &    0.52 (0.09)                                &    0.55 (0.11)                & \textbf{0.69 (0.15)}   \\
                              & Common\_concept           &   0.03 (0.02)   &  0.17 (0.00)   &    0.14 (0.04)                       &    0.09 (0.04)         & \textbf{0.19 (0.05)}    \\
                              & Object\_counting          &    0.30 (0.02)    & \textbf{0.50 (0.06)} &    0.38 (0.06)                             &    0.40 (0.12)      &  0.41 (0.03)\\
                              & Odd\_one\_out             &    0.32 (0.02)  & \textbf{0.64 (0.04)}  &   0.57 (0.02)                              &    0.25 (0.18)        & \textbf{0.64 (0.00)}      \\
                              & Orthography\_starts\_with &    0.23 (0.01)  &  0.47 (0.02)  &    0.41 (0.09)                            & 0.54 (0.06) &      \textbf{0.60 (0.12) }     \\
                              & Taxonomy\_animal          &     0.02 (0.02)  & 0.38 (0.15)  &    0.67 (0.14)                       
                              &    \textbf{0.85 (0.06)}      &    0.71 (0.02) \\
& Auto\_categorization      &   \textbf{0.31 (0.01)}    &  0.20 (0.03) &    0.29 (0.02)                                &    0.07 (0.07)                & 0.29 (0.04)      \\
&Word\_sorting             &    0.58 (0.01)      & 0.01 (0.00)  &  0.64 (0.05)                            &    0.23 (0.20)      &      \textbf{0.70 (0.03)}    \\
    \hline
    \textit{CLUE} & Sentence\_similarity      &   0.00 (0.00)     & 0.05 (0.00) &    0.10 (0.00)                             &    0.00 (0.00)    & \textbf{0.13 (0.07)}  \\
                       & Sentiment                 &  \textbf{0.90 (0.00)}    &  0.63 (0.17)  &    0.88 (0.03)                             &    0.88 (0.02)      &    0.89 (0.01)                          \\
    \hline
    \textit{Translation} 
     &Num\_to\_verbal           &   0.13 (0.02)   & \textbf{1.00 (0.00)}    &   0.99 (0.01)                             &    \textbf{1.00 (0.00)}      &  0.99 (0.01)    \\
    & Translation\_en-de        &    \textbf{0.83 (0.01)}    &  0.80 (0.02)  &    0.82 (0.01)                             &    0.77 (0.02)     &      0.82 (0.01)  \\
                         & Translation\_en-es        & 0.86 (0.01)    &    0.76 (0.00)  &    0.67 (0.24)                             &   \textbf{0.89 (0.00)}   & 0.87 (0.02)   \\
                         & Translation\_en-fr        &   \textbf{0.88 (0.01)}    &  0.86 (0.00)  &   0.77 (0.06)                    
                         &   0.85 (0.02)     & 0.83 (0.01)   \\
 \hline
    \textit{Style}  & Informal\_to\_formal      &   \textbf{0.57 (0.01)}   &  0.50 (0.02)  &   0.48 (0.02)  &    0.54 (0.09)      & 0.44 (0.05)\\
    \hline
    \textit{Coding} 
                       & Auto\_debugging           &   \textbf{0.25 (0.00)}   &  \textbf{0.25 (0.00)}  &    \textbf{0.25 (0.00)}    &    0.07 (0.07)                & \textbf{0.25 (0.00)}   \\

    \hline
    &median score              &       0.49        &  0.63 & 0.66                                    &     0.64            &        \textbf{0.71}        \\    
    &\# best-performing tasks  &      8               &   8  &  8                                     &       7              &       \textbf{14}        \\
    \hline
    \end{tabular}}}
    \end{center}
    \caption{\small Average test performance (and standard deviations) across 3 random seeds comparing \textsc{ACING} versus APE~\citep{zhou2023large}, EvoPrompt~\citep{guo2023connecting}, InstructZero~\citep{instruct}, and INSTINCT~\cite{instinct}. The bottom rows report the median score and total number of best-performing tasks for each method.}
    \label{tab:app_instruction_induction}
\end{table*}

\section{ACING vs InstructZero on GPT-4}
\label{secApp:gpt-4}

We extend our experiments to include GPT-4 as the black-box LLM and Vicuna as the white-box model. Given the high cost associated with querying GPT-4, we restrict our comparison to InstructZero, the strongest baseline after ACING based on prior results with GPT-3.5 (Table~\ref{tab:app_instruction_induction}). To ensure a fair and cost-efficient comparison, both methods are allocated an equal budget of 100 API calls. Additionally, we focus this analysis on the most challenging subset—cognitive tasks—to concentrate the evaluation on settings where prompt optimization is most demanding.

Table~\ref{tab:ACING_gpt4} presents the results of this experiment. Consistent with our earlier findings (Table~\ref{tab:app_instruction_induction}), ACING continues to excel in cognitive tasks, solving 6 out of 8 tasks compared to only 3 solved by InstructZero, and achieving a 13-point higher median score. These results further highlight ACING’s robustness and effectiveness under tighter budgets and more difficult evaluation scenarios.

\begin{table*}[h]
    \begin{center}
    \small
    \resizebox{0.81\textwidth}{!}{
    \begin{tabular}{llcc}
    \hline
    \textbf{Category}          & \textbf{Task}             &  \textbf{InstructZero}   & \textbf{ACING}\\
    \hline
   
    \textit{Cognitive Tasks on GPT-4} & Cause\_and\_effect         &    0.43 (0.10) & \textbf{0.53 (0.10)}   \\
                              & Common\_concept     &  \textbf{0.27 (0.00)}    & 0.04 (0.01)    \\
                              & Object\_counting      &    0.54 (0.03)      &  \textbf{0.62 (0.03)}\\
                              & Odd\_one\_out     &    \textbf{0.77 (0.01)}        & \textbf{0.77 (0.01)}      \\
                              & Orthography\_starts\_with & 0.40 (0.12) &      \textbf{0.60 (0.03) }     \\
                              & Taxonomy\_animal  &    0.97 (0.02)      &    \textbf{0.98 (0.00)} \\
& Auto\_categorization   & \textbf{0.35 (0.01) }               & 0.31 (0.02)      \\
&Word\_sorting           &    0.72 (0.03)      &      \textbf{0.74 (0.01)}    \\
    \hline

    \hline
    &median score          &     0.48           &        \textbf{0.61}        \\    
    &\# best-performing tasks  &  3              &       \textbf{6}        \\
    \hline
    \end{tabular}}
    \end{center}
    \caption{\small Average test performance (with standard deviations) across 3 random seeds comparing \textsc{ACING} and InstructZero \cite{instruct} on 8 cognitively demanding tasks using \textbf{GPT-4} (black-box) and Vicuna (white-box), under a budget of 100 calls. The bottom rows report the median score and the total number of tasks where each method achieves the highest performance.}

    \label{tab:ACING_gpt4}
\end{table*}

\section{Extended Results on GPT-4o}
\label{appendix:gpt4o_results}

We extend a subset of our evaluation to include results on GPT-4o \cite{hurst2024gpt}, as shown in Table~\ref{tab:gpt4o_results}. These results are based on 3 repetitions, and we report the average accuracy along with standard deviation in parentheses. We observe that \textsc{ACING} continues to perform competitively, often outperforming human-written instructions on tasks such as \textit{Antonyms}, \textit{Rhyme}, and \textit{Sentence Similarity}. 

\begin{table}[h]
\centering
\small
\begin{tabular}{@{}lcc@{}}
\toprule
\textbf{Task with GPT-4o} & \textbf{Human} & \textbf{ACING} \\
\midrule
Antonyms              & 0.73 (0.01) & \textbf{0.80} (0.01) \\
Rhyme                 & 0.61 (0.01) & \textbf{0.80} (0.15) \\
Sentence Similarity   & 0.00 (0.02) & \textbf{0.15} (0.00) \\
Informal to Formal    & \textbf{0.58} (0.00) & 0.48 (0.02) \\
Synonyms              & 0.14 (0.00) & \textbf{0.20} (0.05) \\
Cause and Effect      & \textbf{0.92} (0.03) & 0.85 (0.13) \\
\bottomrule
\end{tabular}
\caption{Comparison of human-written (from \cref{tab:human_performance_appendix} vs.\ \textsc{ACING}-generated instructions on a subset of tasks evaluated using GPT-4o \cite{hurst2024gpt}. Results are averaged over 3 repetitions; standard deviations are shown in parentheses.}
\label{tab:gpt4o_results}
\end{table}

\section{Ablation Studies}
\label{app:Ablation_study}
We perform ablation studies to understand the role of key design choices in ACING’s performance. These include critic usage, action dimensionality, exemplar count, budget allocation, and white-box model selection.

\subsection{On the use of Critics}
\label{app:ablation_design}

To empirically assess the contribution of the critic(s), we evaluate three variants of our method:

\begin{itemize}
    \item \textbf{ACING (two critics):} Full method using dual critics.
    \item \textbf{ACING (one critic):} Ablation using a single critic.
    \item \textbf{Policy Gradient (no critic):} Baseline using pure policy gradients without any critic.
\end{itemize}

We report results averaged across 30 diverse tasks (3 trials per task). Due to space constraints, we summarize performance using per-category averages and global statistics.

\noindent
\textit{Summary:} In Table~\ref{tab:acing_vs_reinforce}, ACING achieves a slightly higher median score (0.71 vs. 0.70) and clearly dominates in terms of robustness, outperforming policy gradient on 21 of 33 tasks. This suggests that the critic plays a crucial role in stabilizing learning and guiding exploration, especially in noisy or sparse reward settings (e.g., Rhymes, Auto\_categorization, Word\_sorting). 

Table~\ref{tab:acing_vs_acing_one_critic} further underscores the benefit of using two critics: while both ACING variants reach the same median score (0.71), the two-critic version leads on more tasks (22 vs. 16) and shows stronger performance on several cognitively demanding or unstable tasks (e.g., Cause\_and\_effect, Sentence\_similarity, Taxonomy\_animal). This highlights the importance of ensemble value estimation in improving reliability across diverse tasks.

\begin{table*}[h]
    \begin{center}
    \resizebox{0.8\textwidth}{!}{
    \begin{tabular}{llcc}
    \hline
    \textbf{Category} & \textbf{Task} & \textbf{ACING} & \textbf{Policy Gradient (no critic)}  \\
    \hline
    \textit{Spelling} & Letters\_list & \textbf{1.00 (0.00)} & 0.93 (0.05) \\
    & First\_word\_letter & \textbf{1.00 (0.00)} & \textbf{1.00 (0.00)} \\
    & Second\_word\_letter &\textbf{ 0.70 (0.15)} & 0.55 (0.30)  \\
    \hline
    \textit{Morpho-Syntax} & Singular\_to\_plural & 0.95 (0.03) & \textbf{1.00 (0.00) }\\
    & Active\_to\_passive &\textbf{ 1.00 (0.00)} & \textbf{1.00 (0.00)}  \\
    & Negation & \textbf{0.71 (0.06)} & \textbf{0.71 (0.00)}  \\
    \hline
    \textit{Lexical Semantics} & Antonyms & \textbf{0.74 (0.01)} & 0.69 (0.15) \\
    & Synonyms & 0.13 (0.02) & \textbf{0.19 (0.10)} \\
    & Word\_unscrambling & 0.50 (0.07) &\textbf{ 0.54 (0.02)} \\
    \hline
    \textit{Phonetics} & Rhymes & \textbf{0.57 (0.31)} & 0.36 (0.14) \\
    \hline
    \textit{Numerical} & Sum & \textbf{1.00 (0.00)} & 0.98 (0.03) \\
    & Diff & \textbf{1.00 (0.00) }& 0.93 (0.05)\\
    \hline
    \textit{Knowledge} & Larger\_animal & \textbf{0.84 (0.07)} & 0.68 (0.18) \\
    & Periodic\_elements & \textbf{0.98 (0.00)} & 0.93 (0.04)\\
    \hline
    \textit{Cognitive Tasks} & Cause\_and\_effect & 0.69 (0.15) & \textbf{0.71 (0.19)}\\
    & Common\_concept &\textbf{ 0.19 (0.05)} & 0.12 (0.02)\\
    & Object\_counting & 0.41 (0.03) & 0.44 (0.08)\\
    & Odd\_one\_out &\textbf{ 0.64 (0.00) }& 0.53 (0.08)\\
    & Orthography\_starts\_with & \textbf{0.60 (0.12)} & 0.52 (0.14)\\
    & Taxonomy\_animal & 0.71 (0.02) & \textbf{0.77 (0.10)} \\
    & Auto\_categorization &\textbf{ 0.29 (0.04)} & 0.17 (0.12) \\
    & Word\_sorting & \textbf{0.70 (0.03)} & 0.23 (0.30)  \\
    \hline
    \textit{CLUE} & Sentence\_similarity &  \textbf{0.13 (0.07)} & 0.00 (0.00) \\
    & Sentiment & \textbf{0.89 (0.01)} & \textbf{0.89 (0.00)} \\
    \hline
    \textit{Translation} 
    & Num\_to\_verbal & 0.99 (0.01) & \textbf{1.00 (0.00)}\\
    & Translation\_en-de & \textbf{0.82 (0.01)} & 0.80 (0.00)\\
    & Translation\_en-es &\textbf{ 0.87 (0.02)} & \textbf{0.87 (0.01)} \\
    & Translation\_en-fr & 0.83 (0.01) & \textbf{0.87 (0.00)}\\
    \hline
    \textit{Style} & Informal\_to\_formal & 0.44 (0.05) & \textbf{0.50 (0.03)}\\
    \hline
    \textit{Coding} & Auto\_debugging & \textbf{0.25 (0.00)} & \textbf{0.25 (0.00)}\\
    \hline
     &median score    &       \textbf{ 0.71 }    &     0.70      \\    
    &\# best-performing tasks  & \textbf{21} &   14        \\
    \hline
    \end{tabular}}
    \end{center}
    \caption{Performance comparison between ACING and Reinforce different task categories.}
    \label{tab:acing_vs_reinforce}
\end{table*}

\begin{table*}[h]
    \begin{center}
    \resizebox{0.8\textwidth}{!}{
    \begin{tabular}{llcc}
    \hline
    \textbf{Category} & \textbf{Task} & \textbf{ACING (two critics)} &\textbf{ACING (one critic)} \\
    \hline
    \textit{Spelling} & Letters\_list &\textbf{ 1.00 (0.00)}  & \textbf{1.00 (0.00)} \\
    & First\_word\_letter & \textbf{1.00 (0.00)} & \textbf{1.00 (0.00)} \\
    & Second\_word\_letter & \textbf{0.70 (0.15)} & 0.58 (0.31) \\
    \hline
    \textit{Morpho-Syntax} & Singular\_to\_plural & 0.95 (0.03) & \textbf{0.99 (0.00)} \\
    & Active\_to\_passive & \textbf{1.00 (0.00)} & \textbf{1.00 (0.00)} \\
    & Negation & 0.71 (0.06) & \textbf{0.78 (0.01)} \\
    \hline
    \textit{Lexical Semantics} & Antonyms & \textbf{0.74 (0.01)} & 0.72 (0.05) \\
    & Synonyms &\textbf{ 0.13 (0.02) }& \textbf{0.13 (0.00)} \\
    & Word\_unscrambling & 0.50 (0.07)  & \textbf{0.52 (0.04)} \\
    \hline
    \textit{Phonetics} & Rhymes & 0.57 (0.31) &  \textbf{0.63 (0.23)} \\
    \hline
    \textit{Numerical} & Sum & \textbf{1.00 (0.00)} & 0.98 (0.01) \\
    & Diff & \textbf{1.00 (0.00)} & 0.99 (0.02) \\
    \hline
    \textit{Knowledge} & Larger\_animal & \textbf{0.84 (0.07)} & 0.80 (0.12) \\
    & Periodic\_elements & \textbf{0.98 (0.00)}  & 0.97 (0.02) \\
    \hline
    \textit{Cognitive Tasks} & Cause\_and\_effect & \textbf{0.69 (0.15)} & 0.56 (0.00) \\
    & Common\_concept & \textbf{0.19 (0.05)} & 0.16 (0.09) \\
    & Object\_counting & 0.41 (0.03) & \textbf{0.44 (0.09)} \\
    & Odd\_one\_out & \textbf{0.64 (0.00)} & 0.59 (0.05) \\
    & Orthography\_starts\_with & 0.60 (0.12) &  \textbf{0.63 (0.12)} \\
    & Taxonomy\_animal &\textbf{ 0.71 (0.02)} & 0.59 (0.40) \\
    & Auto\_categorization & \textbf{0.29 (0.04)} & \textbf{0.29 (0.06)} \\
    & Word\_sorting & \textbf{0.70 (0.03)} & \textbf{0.70 (0.01)} \\
    \hline
    \textit{CLUE} & Sentence\_similarity &  \textbf{0.13 (0.07)} & 0.05 (0.05) \\
    & Sentiment & \textbf{0.89 (0.01)} & 0.88 (0.02) \\
    \hline
    \textit{Translation} 
    & Num\_to\_verbal & 0.99 (0.01)  &\textbf{ 1.00 (0.00)} \\
    & Translation\_en-de & \textbf{0.82 (0.01)} & 0.81 (0.01) \\
    & Translation\_en-es & \textbf{0.87 (0.02)}  & \textbf{0.87 (0.02)} \\
    & Translation\_en-fr & \textbf{0.83 (0.01)} & 0.82 (0.02) \\
    \hline
    \textit{Style} & Informal\_to\_formal & 0.44 (0.05) & \textbf{0.49 (0.05)} \\
    \hline
    \textit{Coding} & Auto\_debugging & \textbf{0.25 (0.00)} & \textbf{0.25 (0.00)} \\
    \hline
    \hline
     &median score    &       \textbf{ 0.71 }    &     \textbf{0.71 }     \\    
    &\# best-performing tasks  & \textbf{22} &   16        \\
    \hline
    \end{tabular}}
    \end{center}
    \caption{Performance comparison between ACING, and ACING\_one\_critic across different task categories.}
    \label{tab:acing_vs_acing_one_critic}
\end{table*}

\subsection{On Direct Parameterization of the Actor}
\label{app:ablation_actor}

To assess the impact of using a feed-forward network (FFN) versus direct parameterization for the actor, we compare the following two variants:

\begin{itemize}
\item \textbf{ACING:} Our standard approach, where the actor is parameterized using a feed-forward neural network that receives a constant one-dimensional input (set to 1).
\item \textbf{ACING (direct parameterization):} A variant using the same critics, but with the actor represented directly by learnable mean and variance vectors, completely removing the FFN.
\end{itemize}

We report results averaged across 30 diverse tasks (3 trials per task), summarizing performance by task category and global statistics.

\noindent
\textit{Summary:} As shown in Table~\ref{tab:acing_vs_parametrized}, the FFN-based ACING achieves a substantially higher median score (0.69 vs. 0.44) and outperforms the direct parameterization variant on many more tasks (19 vs. 8). While direct parameterization occasionally matches or surpasses FFN performance on simpler tasks (e.g., Negation, Antonyms, Num\_to\_verbal, Informal\_to\_formal), it generally struggles on tasks requiring more compositional or fine-grained control (e.g., Word\_sorting, Odd\_one\_out, Auto\_categorization).

\begin{table*}[h]
    \begin{center}
    \resizebox{0.8\textwidth}{!}{
    \begin{tabular}{llcc}
    \hline
    \textbf{Category} & \textbf{Task} & \textbf{ACING} & \textbf{ACING (w/o FFN)}  \\
    \hline
    \textit{Spelling} & Letters\_list & \textbf{1.00 (0.00)} & 0.99 (0.00) \\
    & First\_word\_letter & \textbf{1.00 (0.00)} & 0.97 (0.03) \\
    & Second\_word\_letter &\textbf{0.70 (0.15)} & 0.21 (0.51)  \\
    \hline
    & Negation & 0.71 (0.06) & \textbf{0.81 (0.03)}  \\
    \hline
    \textit{Lexical Semantics} & Antonyms & 0.74 (0.01) & \textbf{0.75 (0.06)} \\
    & Synonyms & 0.13 (0.02) & \textbf{0.25 (0.09)} \\
    & Word\_unscrambling & \textbf{0.50 (0.07)} & 0.41 (0.06) \\
    \hline
    \textit{Phonetics} & Rhymes & \textbf{0.57 (0.31)} & 0.38 (0.14) \\
    \hline
    \textit{Numerical} & Sum & \textbf{1.00 (0.00)} & 0.79 (0.15) \\
    & Diff & \textbf{1.00 (0.00) }& 0.86 (0.12)\\
    \hline
    \textit{Knowledge} & Larger\_animal & \textbf{0.84 (0.07)} & 0.76 (0.05) \\
    \hline
    \textit{Cognitive Tasks} & Cause\_and\_effect & \textbf{0.69 (0.15)} & 0.53 (0.07)\\
    & Common\_concept &\textbf{0.19 (0.05)} & 0.11 (0.01)\\
    & Object\_counting & \textbf{0.41 (0.03)} & 0.33 (0.08)\\
    & Odd\_one\_out &\textbf{0.64 (0.00)}& 0.37 (0.20)\\
    & Orthography\_starts\_with & \textbf{0.60 (0.12)} & 0.39 (0.17)\\
    & Taxonomy\_animal & \textbf{0.71 (0.02)} & 0.67 (0.06) \\
    & Auto\_categorization &\textbf{0.29 (0.04)} & 0.11 (0.15) \\
    & Word\_sorting & \textbf{0.70 (0.03)} & 0.19 (0.26)  \\
    \hline
    \textit{CLUE} & Sentence\_similarity &  \textbf{0.13 (0.07)} & 0.00 (0.00) \\
    \hline
    \textit{Translation} 
    & Num\_to\_verbal & 0.99 (0.01) & \textbf{1.00 (0.00)}\\
    & Translation\_en-es &\textbf{ 0.87 (0.02)} & \textbf{0.87 (0.01)} \\
    \hline
    \textit{Style} & Informal\_to\_formal & 0.44 (0.05) & \textbf{0.48 (0.03)}\\
    \hline
    \textit{Coding} & Auto\_debugging & \textbf{0.25 (0.00)} & \textbf{0.25 (0.00)}\\
    \hline
     &median score    &       \textbf{ 0.69 }    &     0.44      \\    
    &\# best-performing tasks  & \textbf{19} &   7        \\
    \hline
    \end{tabular}}
    \end{center}
    \caption{Performance comparison between ACING and a variant using the same critics but with direct parameterization of the actor (learnable mean and variance vectors) instead of a feed-forward network.}
    \label{tab:acing_vs_parametrized}
\end{table*}

\subsection{ACING Rewards over the (Calls) Steps}
\label{acing_rewards}

\begin{figure*}[htbp!]
    \centering
    \begin{tabular}{ccc}
        \includegraphics[width=0.3\textwidth, height=3.5cm]{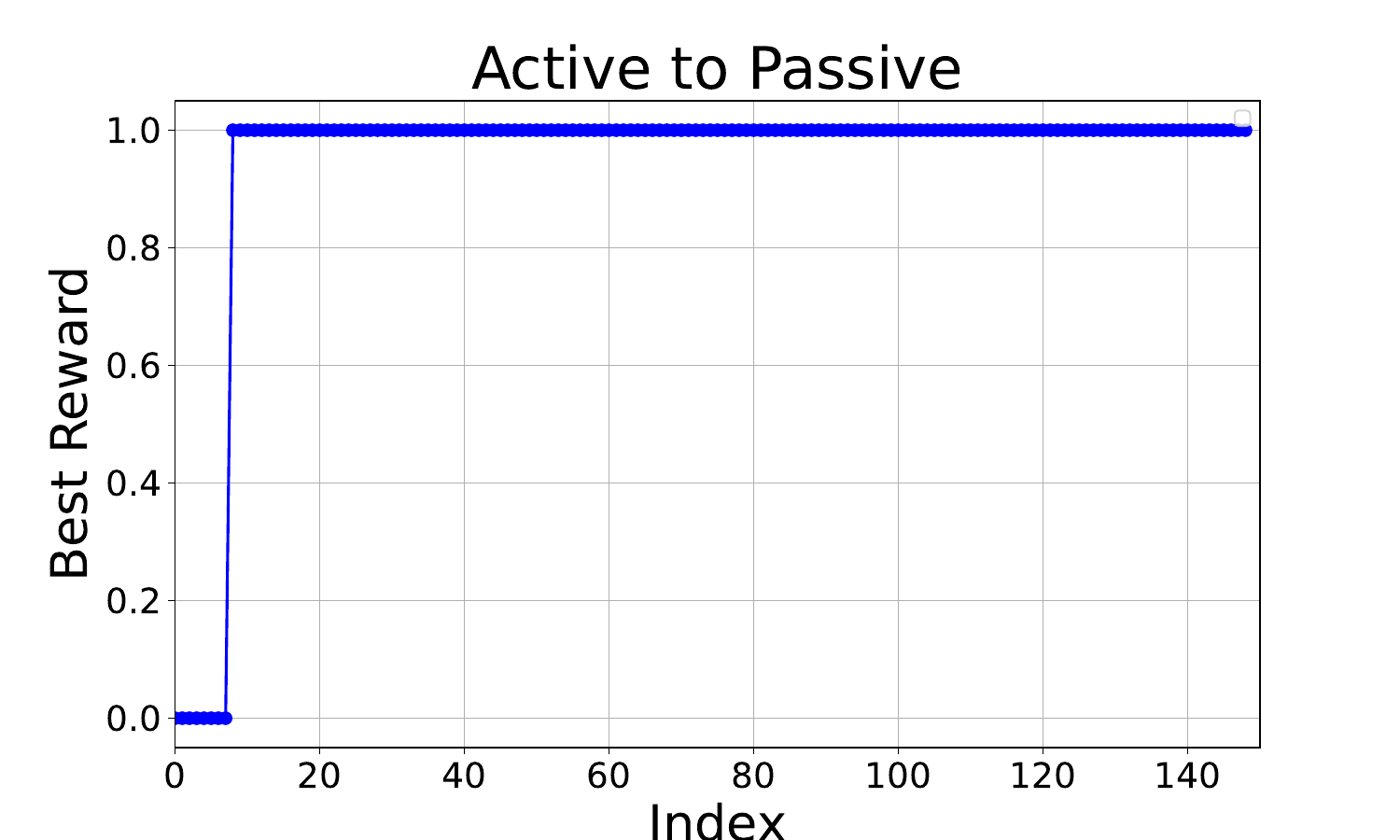} & \includegraphics[width=0.3\textwidth, height=3.5cm]{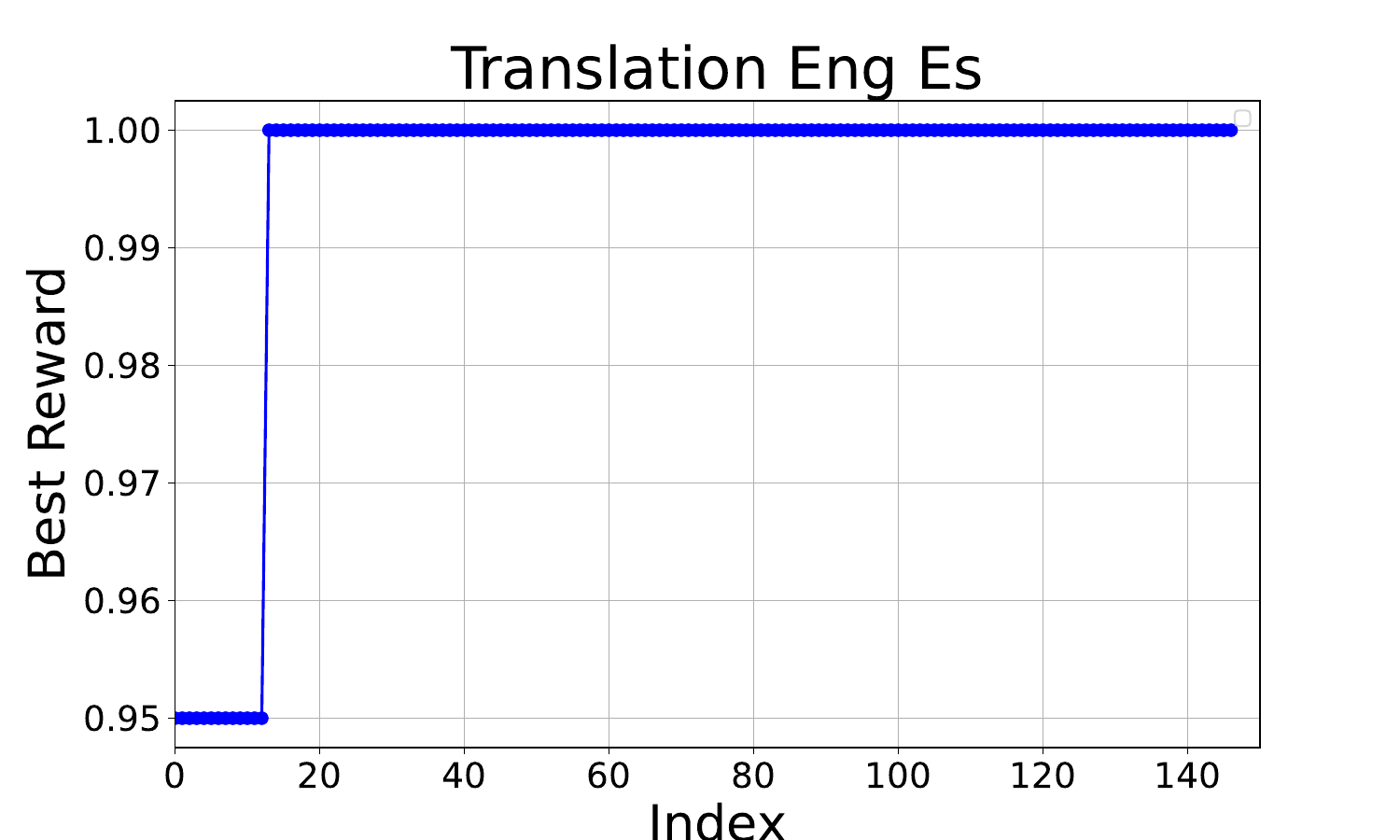} & \includegraphics[width=0.3\textwidth, height=3.5cm]{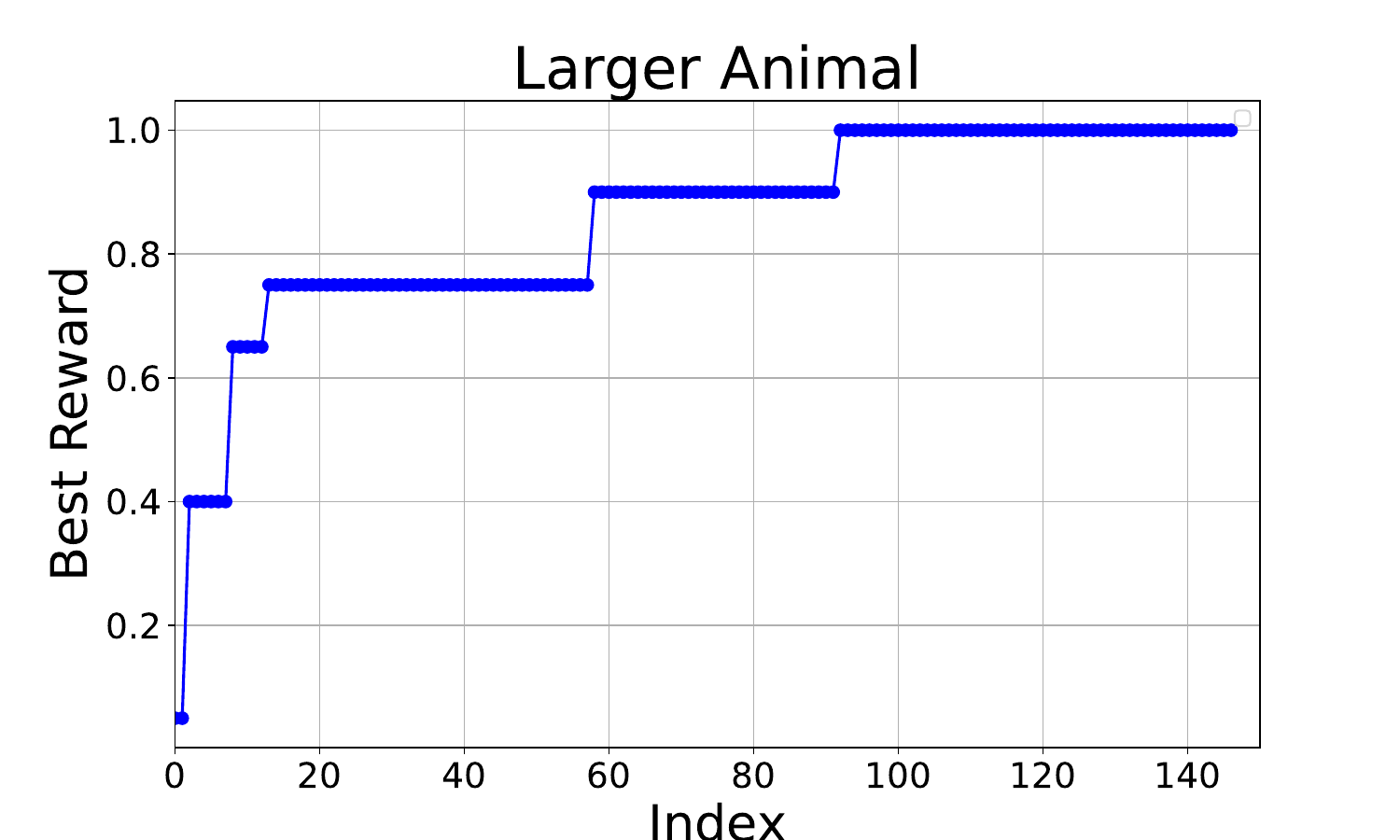} \\
        \includegraphics[width=0.3\textwidth, height=3.5cm]{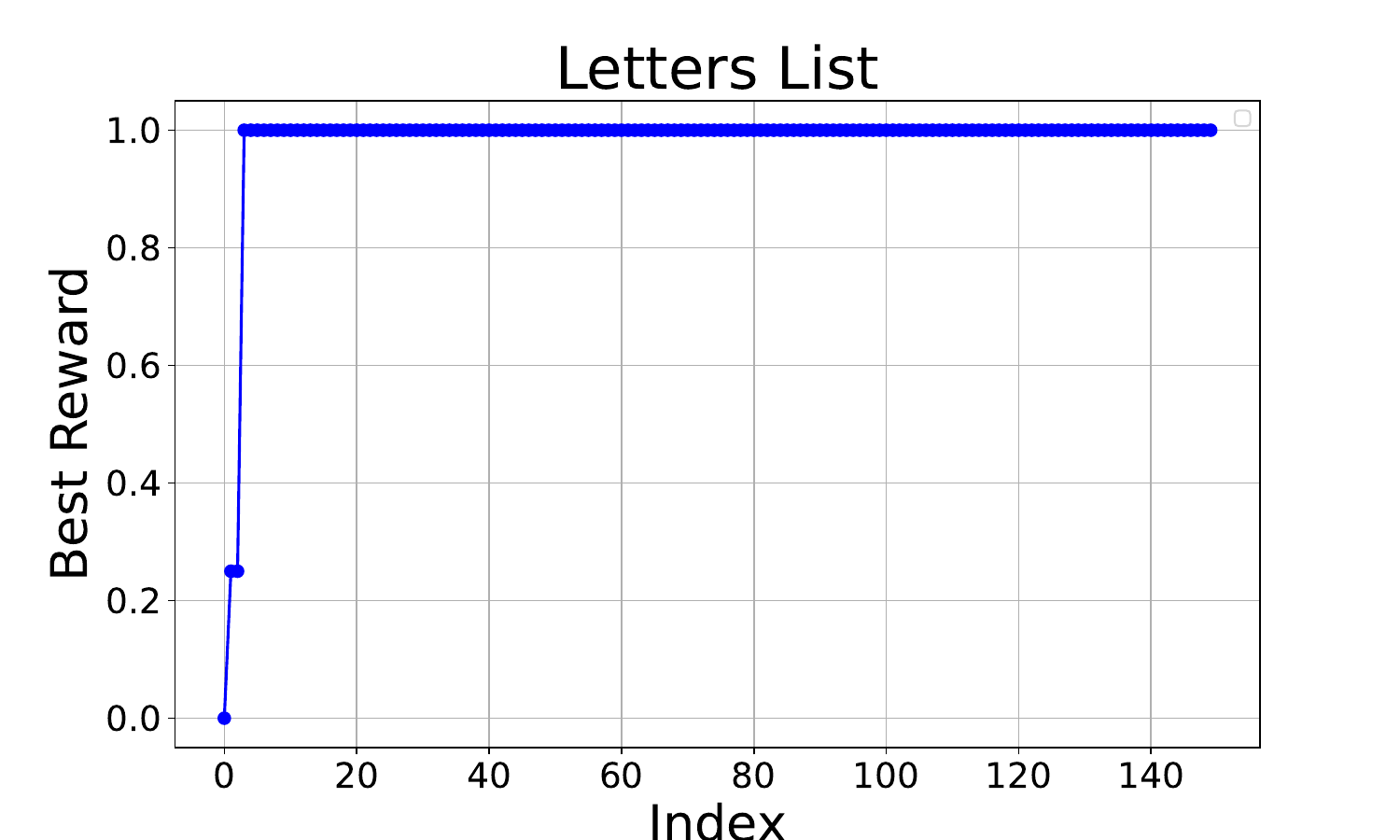} & \includegraphics[width=0.3\textwidth, height=3.5cm]{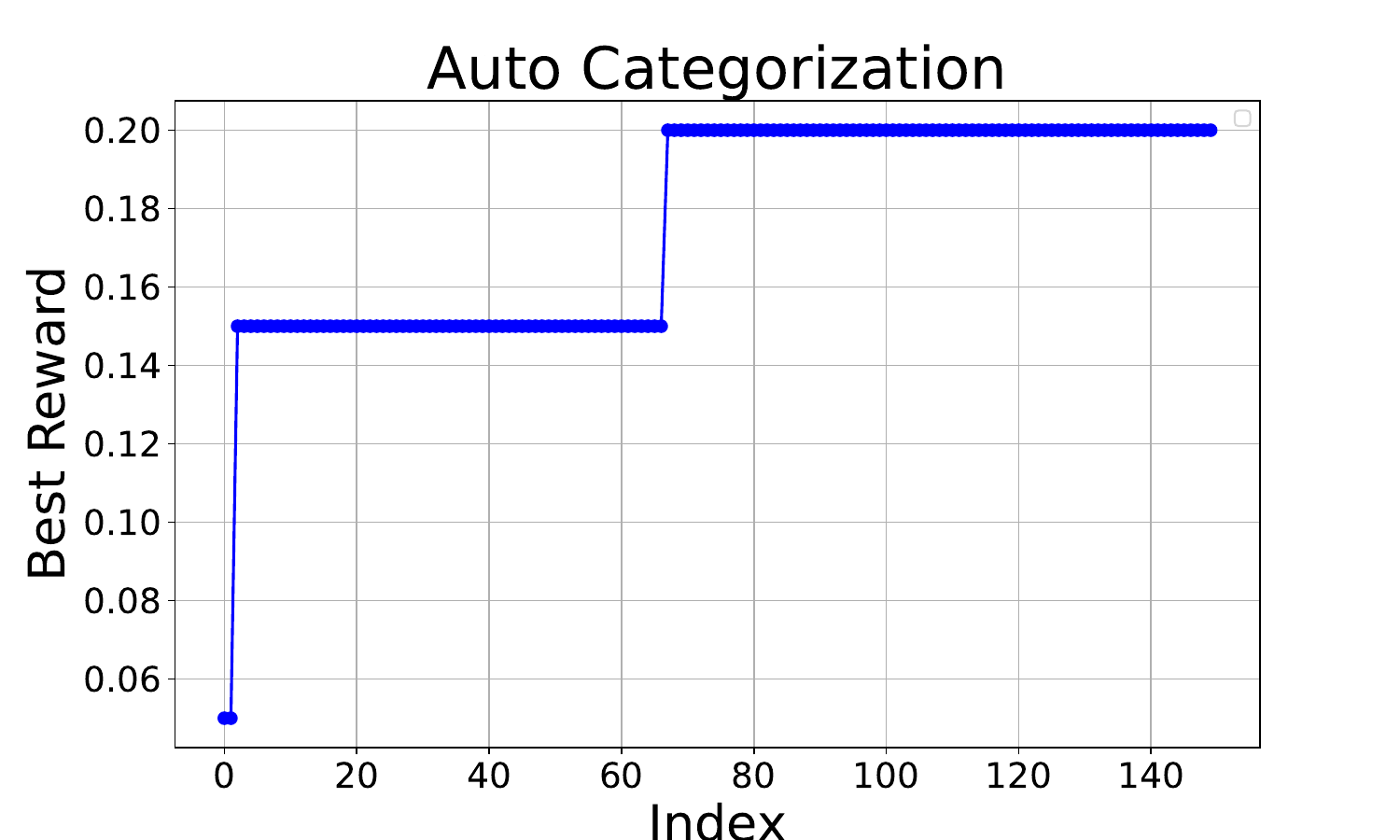} & \includegraphics[width=0.3\textwidth, height=3.5cm]{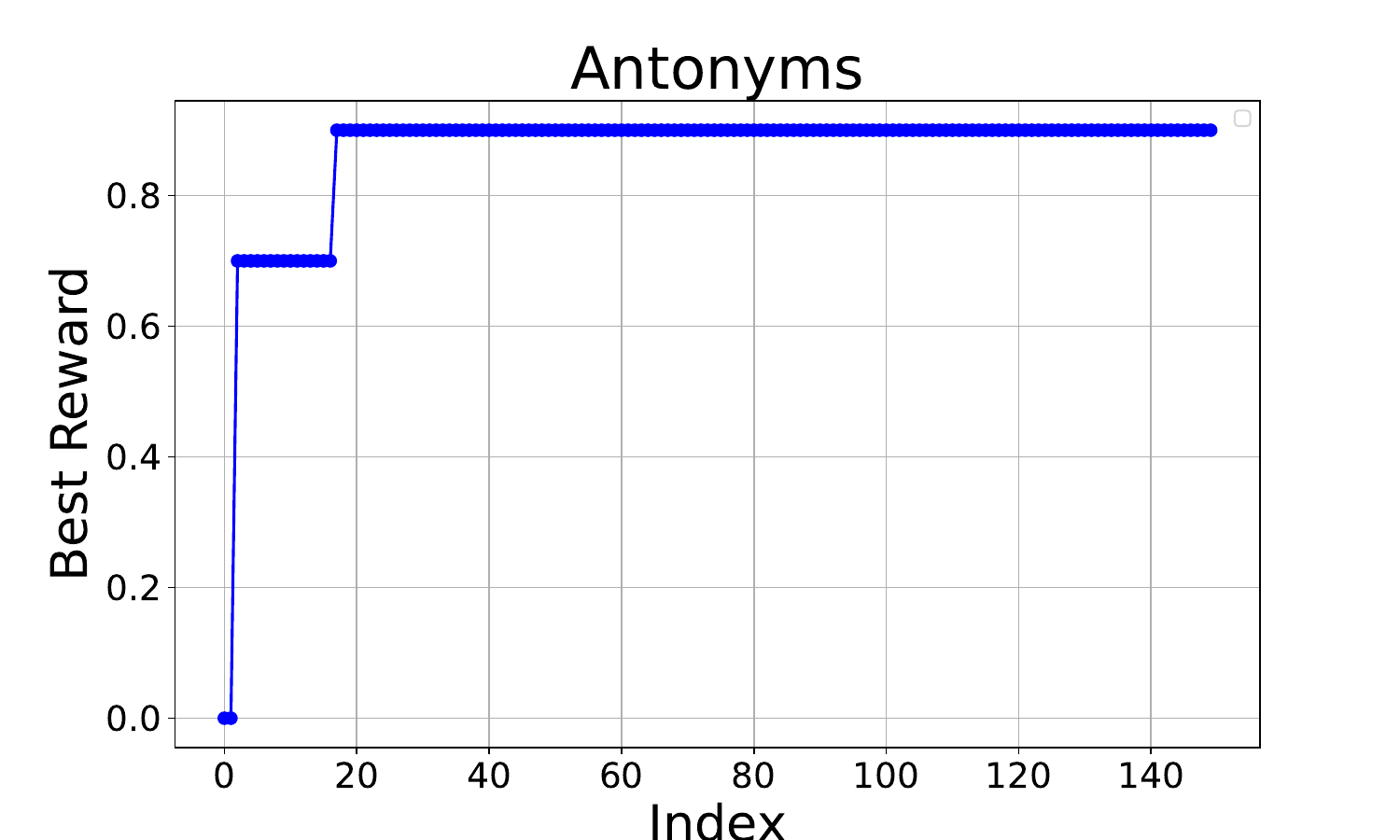} \\
        \includegraphics[width=0.3\textwidth, height=3.5cm]{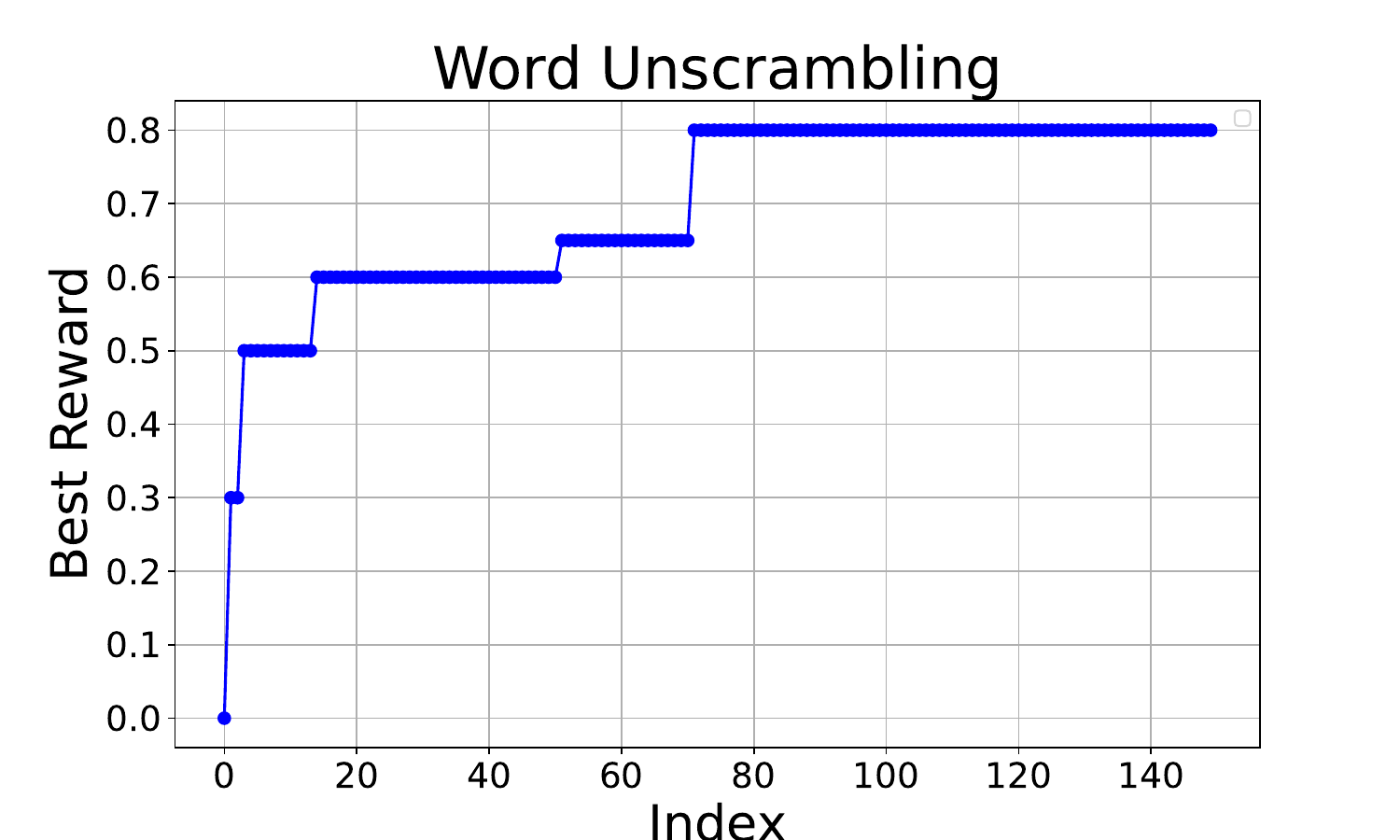} & \includegraphics[width=0.3\textwidth, height=3.5cm]{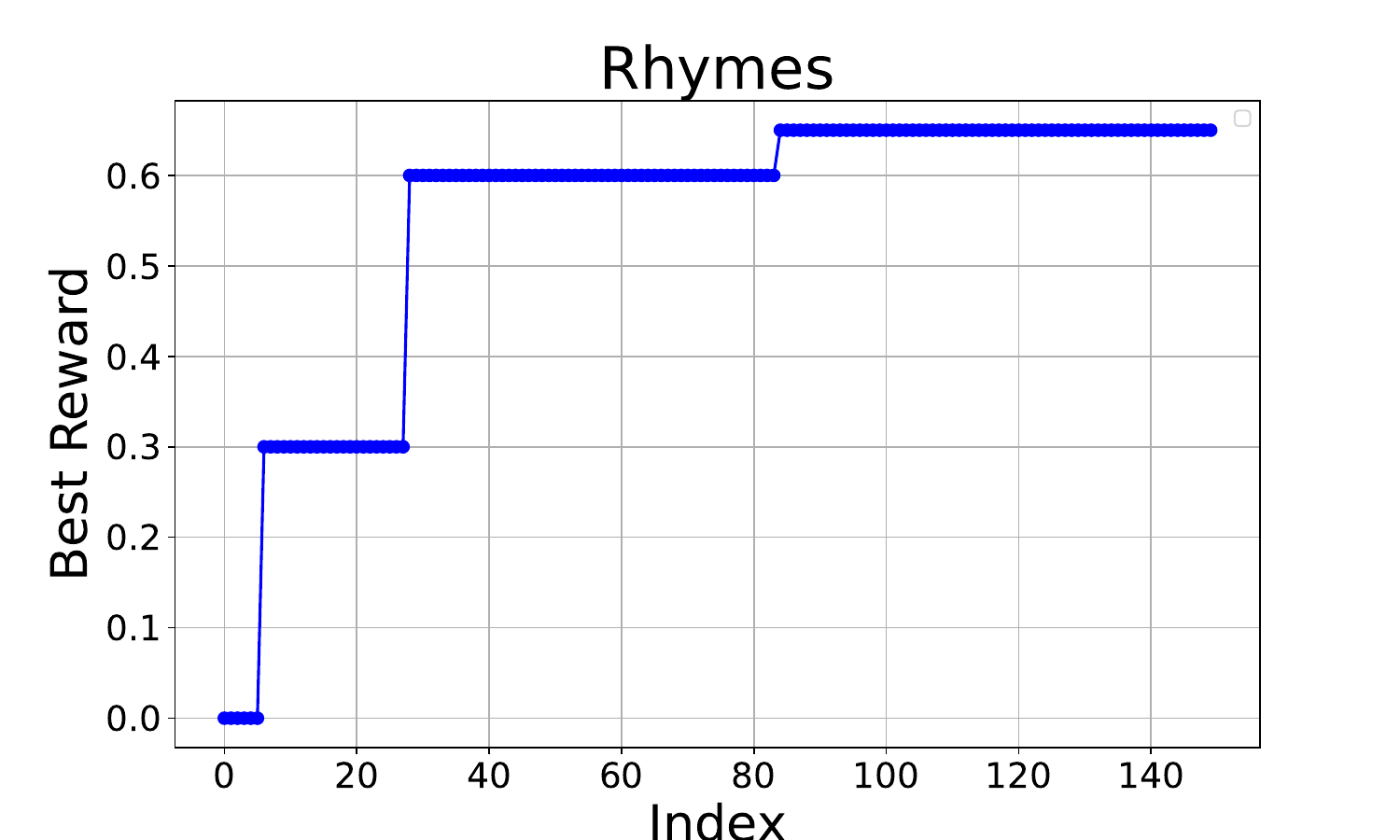} & \includegraphics[width=0.3\textwidth, height=3.5cm]{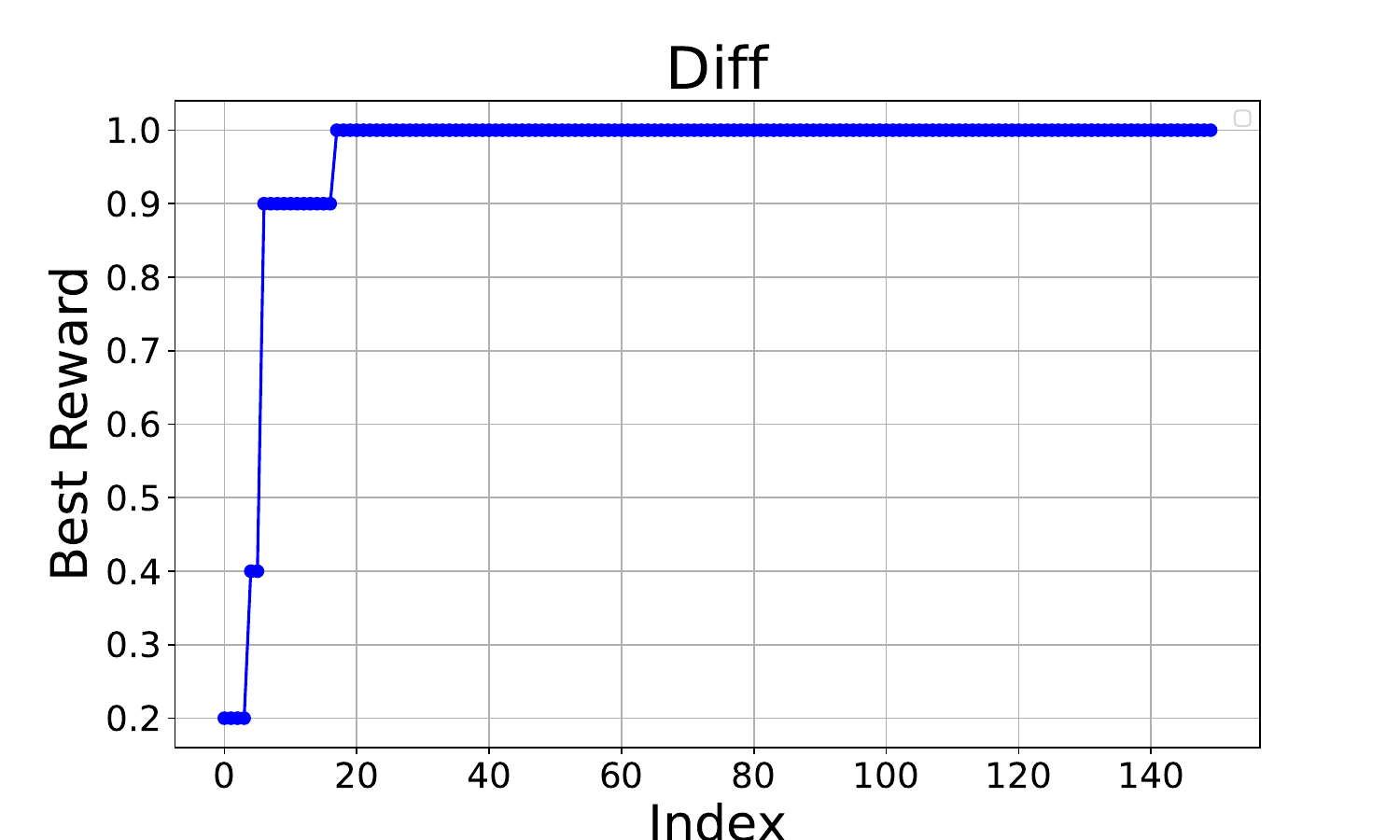} \\
        \includegraphics[width=0.3\textwidth, height=3.5cm]{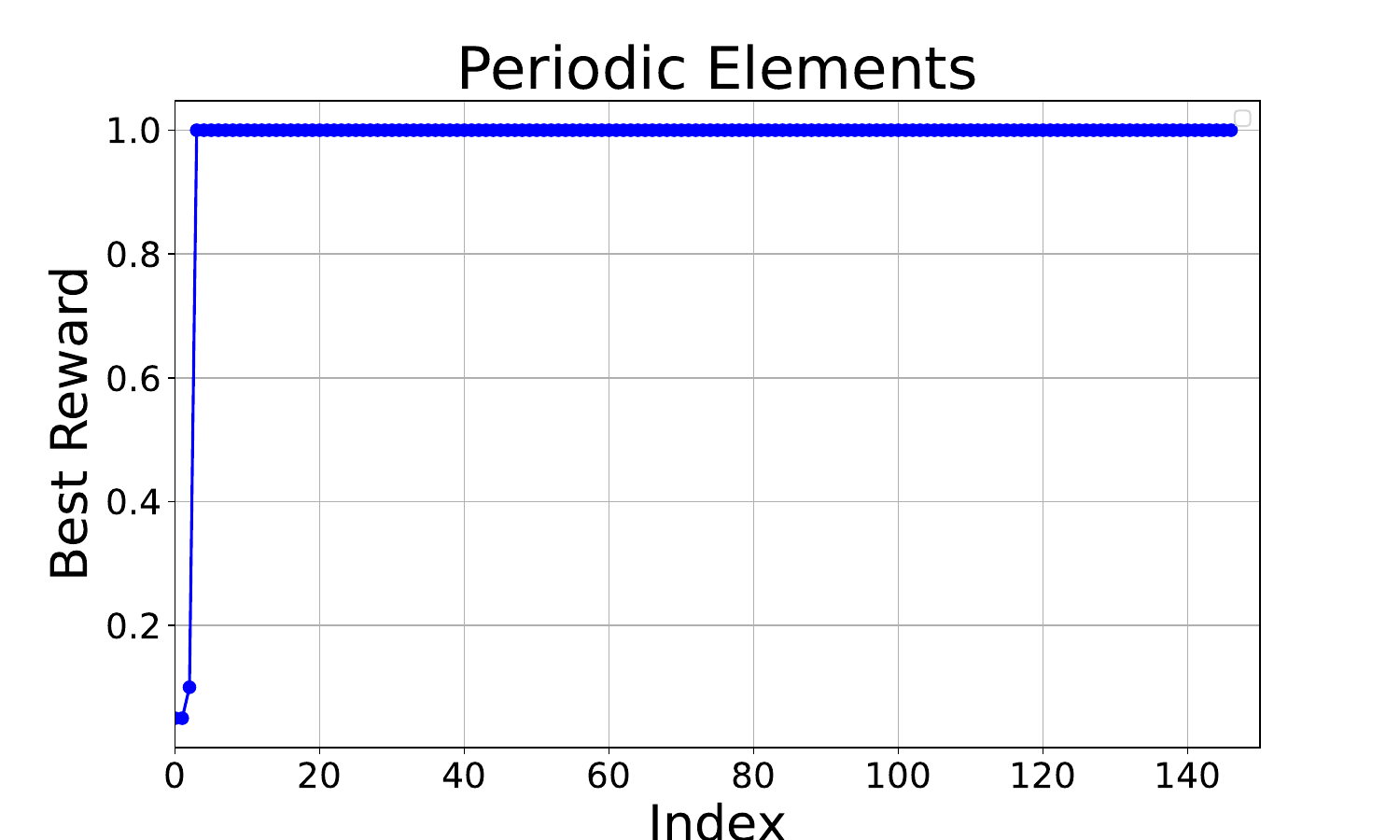} & \includegraphics[width=0.3\textwidth, height=3.5cm]{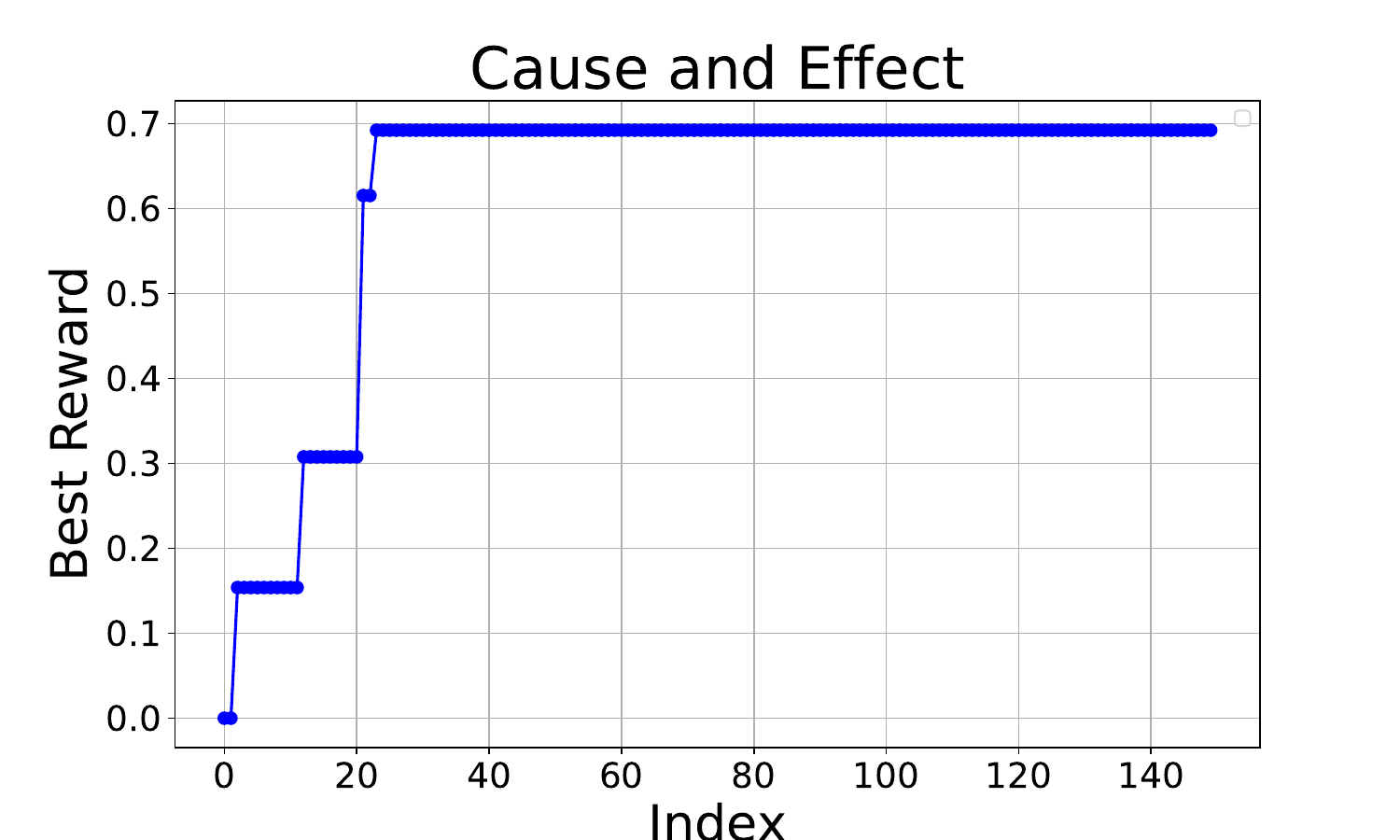} & \includegraphics[width=0.3\textwidth, height=3.5cm]{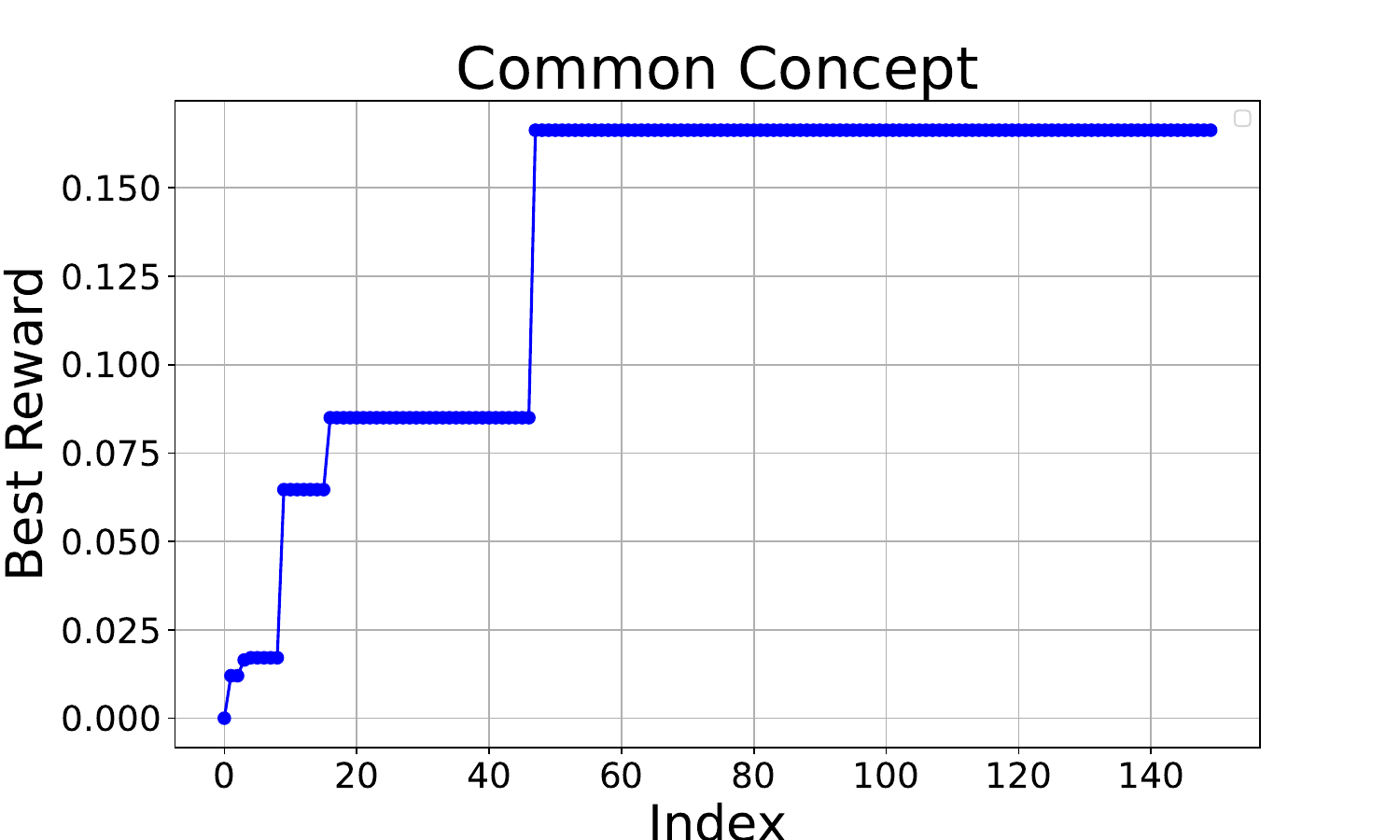} \\
        \includegraphics[width=0.3\textwidth, height=3.5cm]{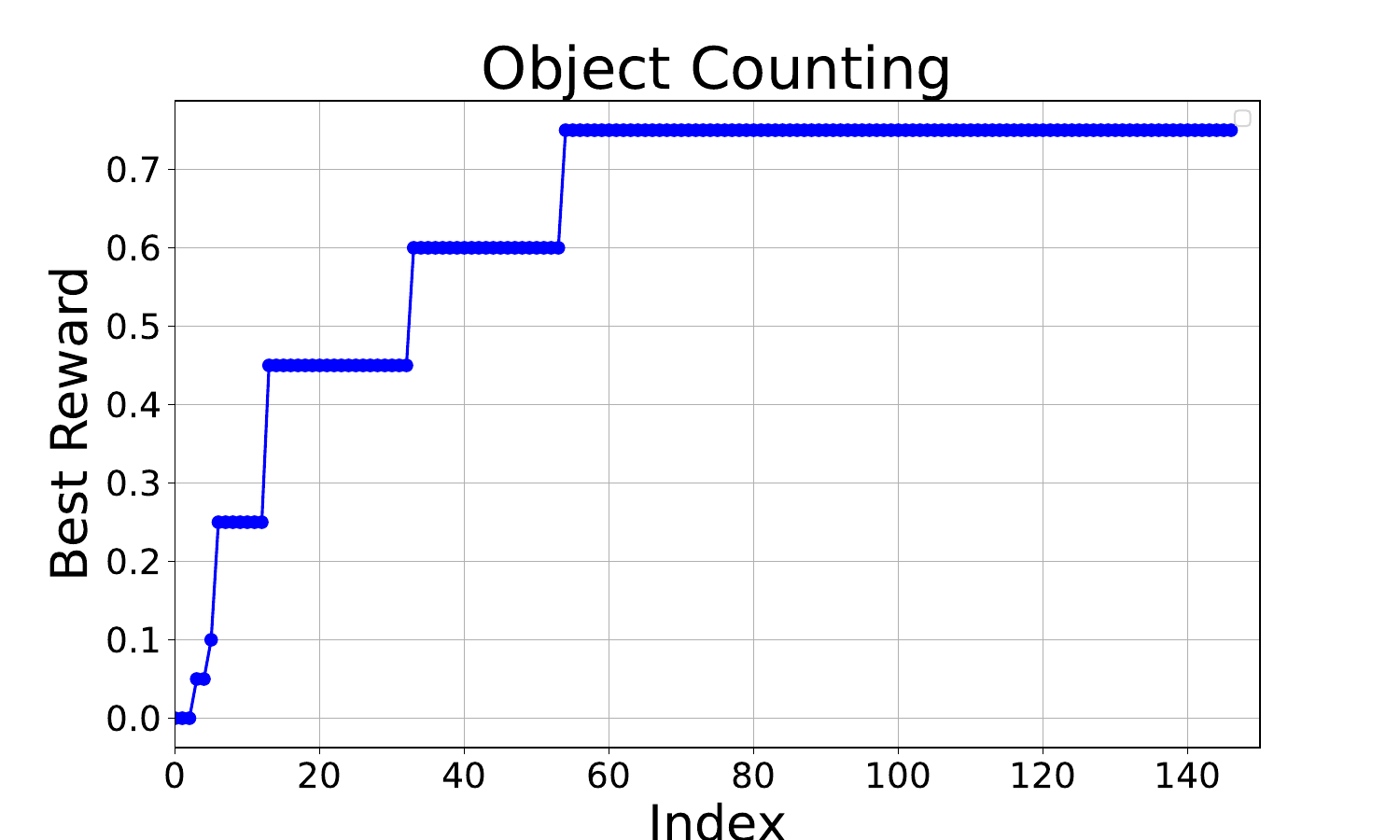} & \includegraphics[width=0.3\textwidth, height=3.5cm]{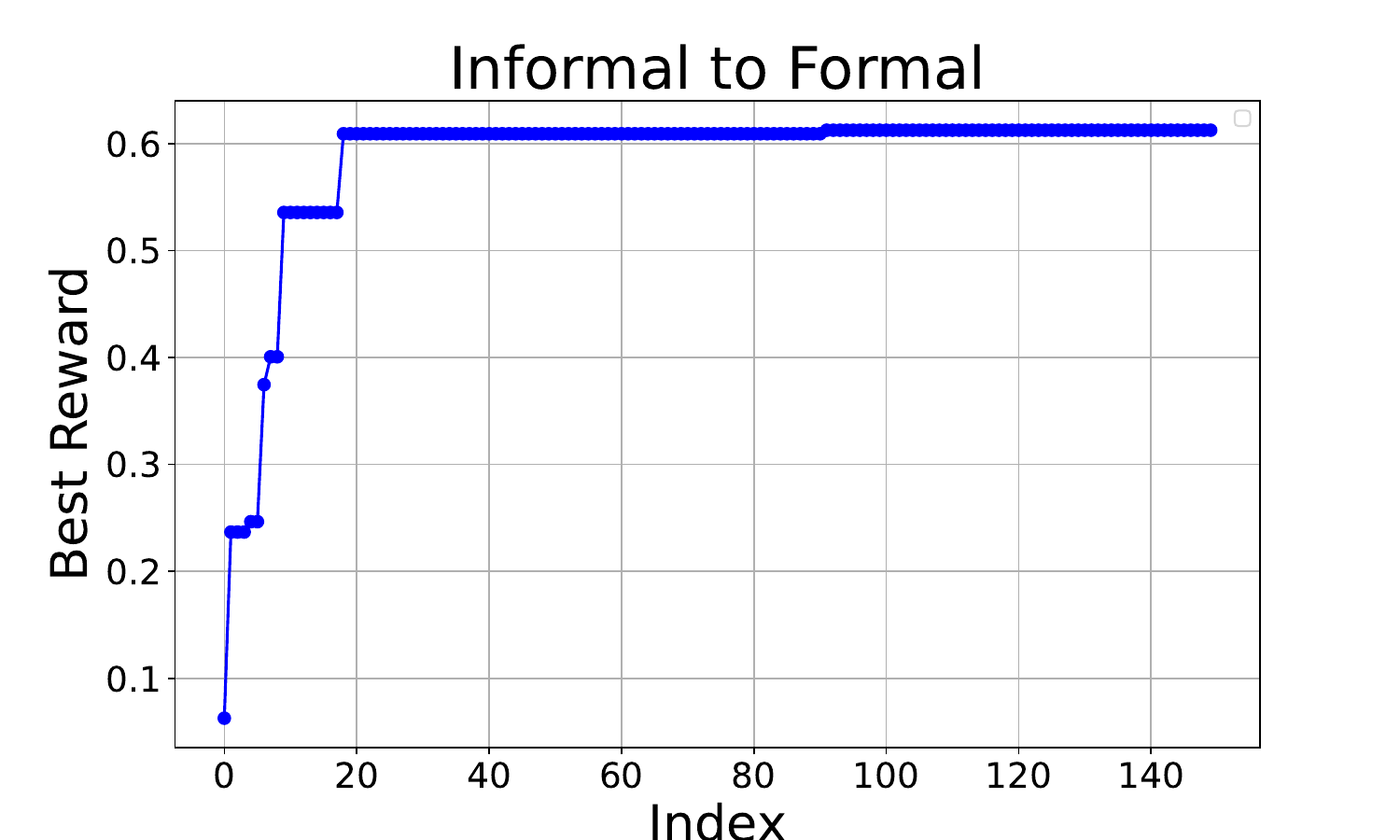} & \includegraphics[width=0.3\textwidth, height=3.5cm]{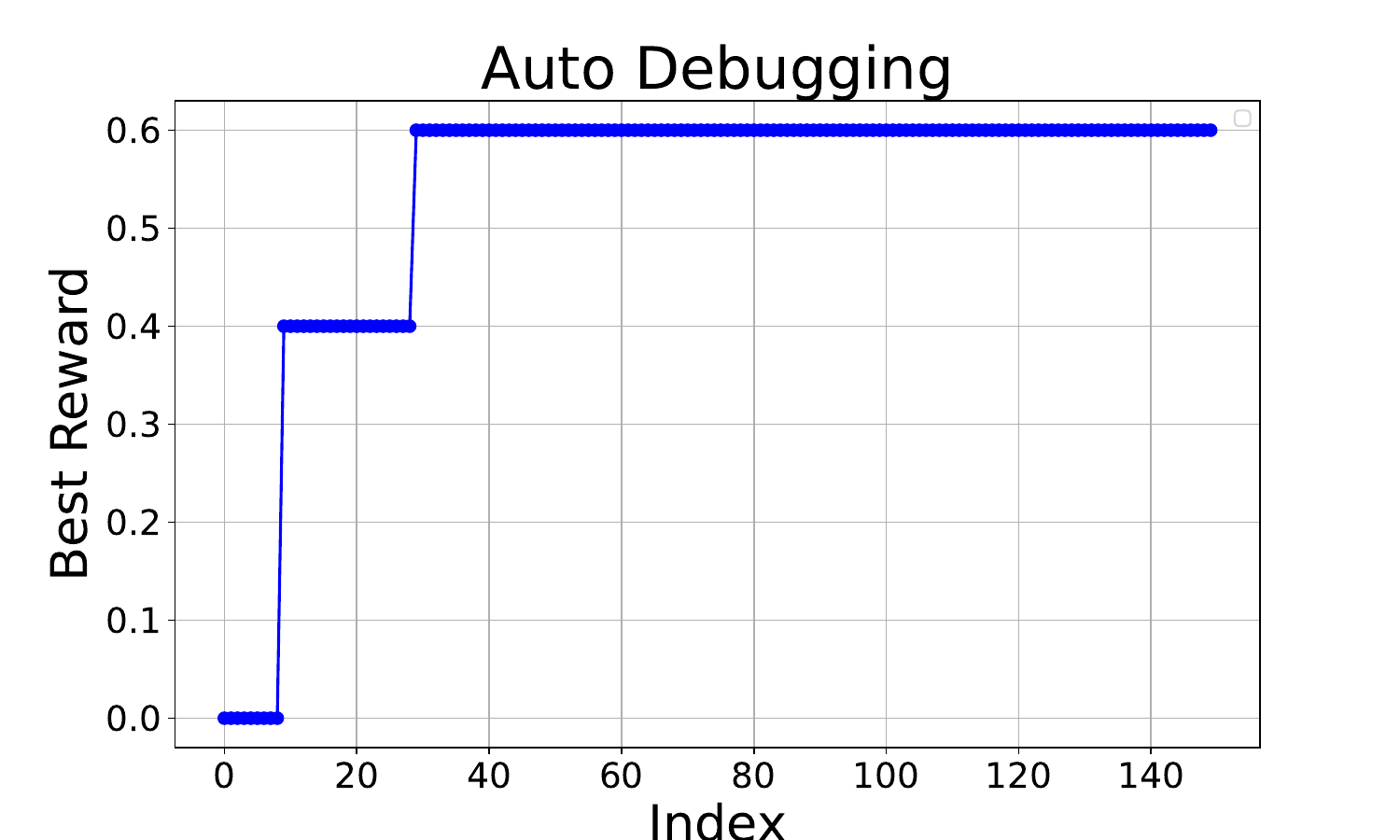} \\
        \includegraphics[width=0.3\textwidth, height=3.5cm]{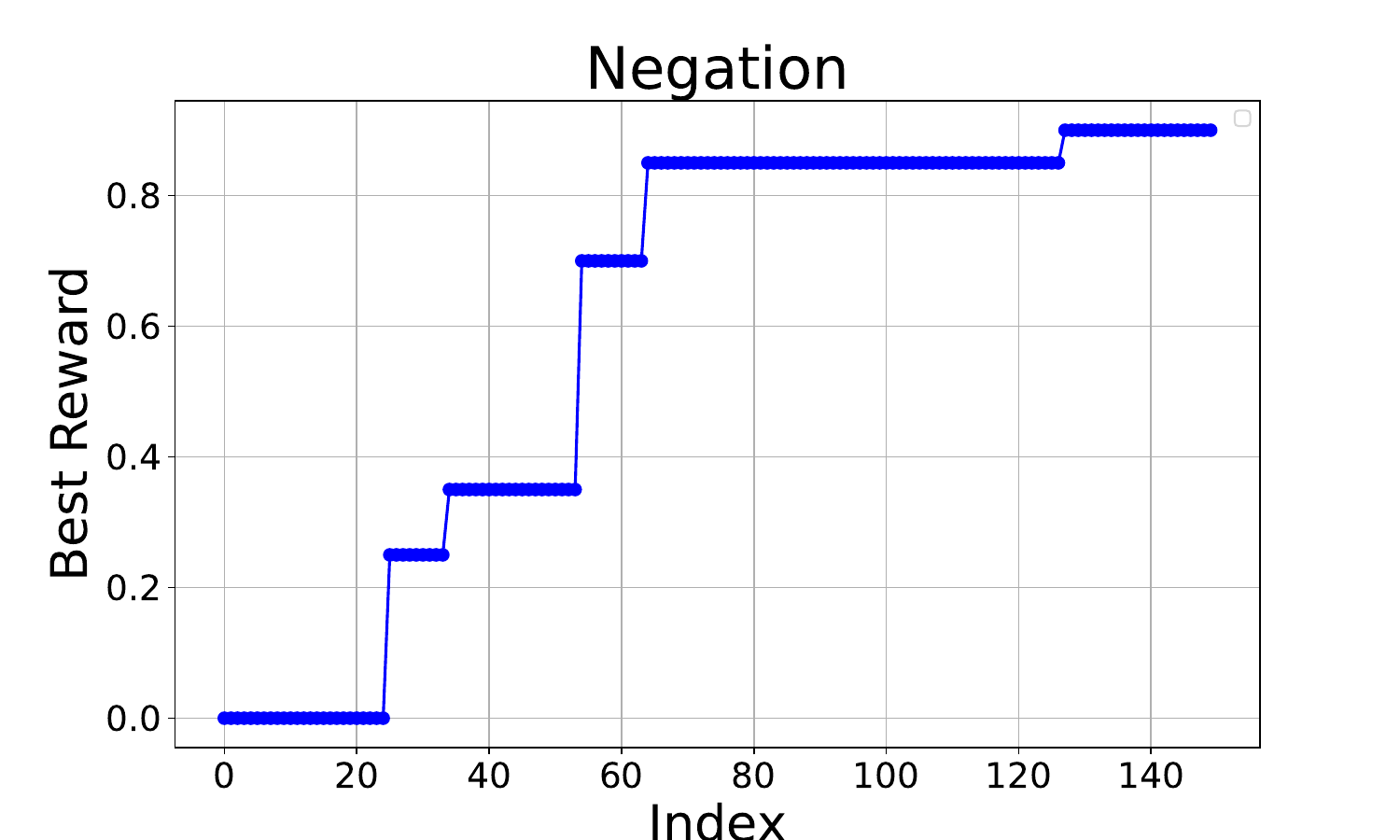} & \includegraphics[width=0.3\textwidth, height=3.5cm]{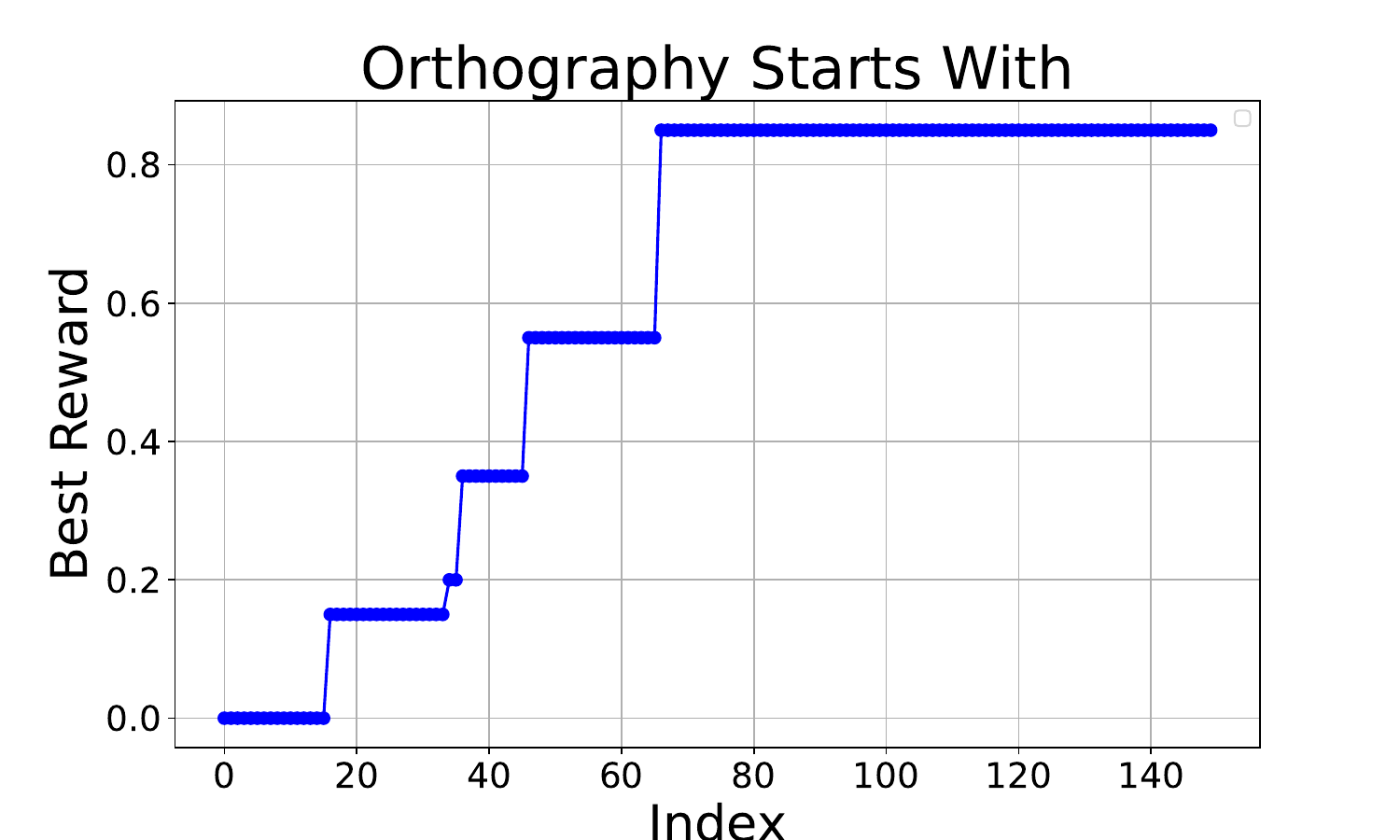} & \includegraphics[width=0.3\textwidth, height=3.5cm]{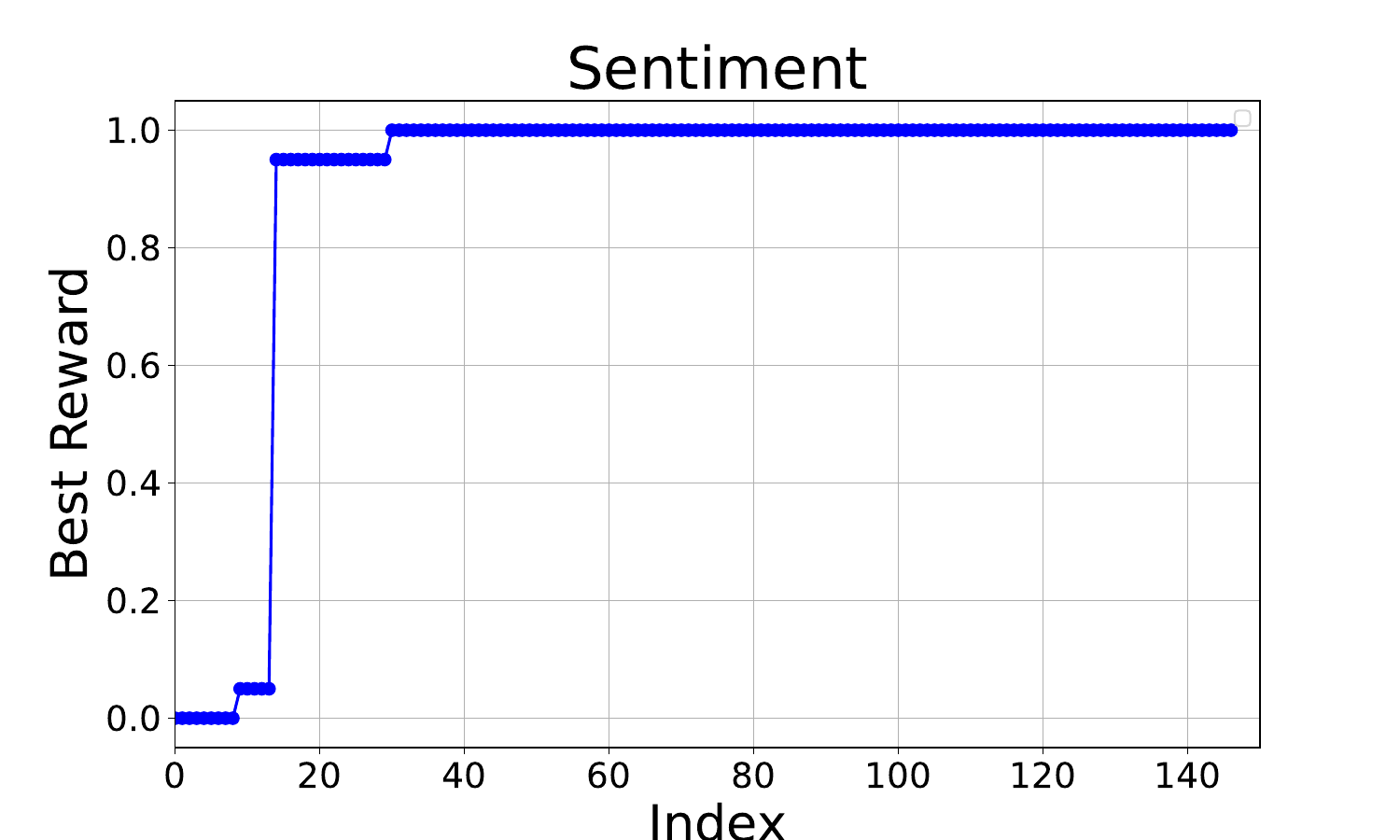} \\
    \end{tabular}
    \caption{Reward plots for running ACING on various selected tasks, showing the highest achieved reward on the y-axis until each API call (step), with the x-axis representing the number of API calls.}
    \label{fig:5x3_grid}
\end{figure*}

In the main paper, we report the final test score after a fixed budget of 165 black-box LLM calls. In this section, we provide reward plots for the ACING approach, showing the best-achieved reward within the conducted calls. As shown in various plots in Figure \ref{fig:5x3_grid}, the ACING approach found the optimal prompt (achieving a reward of 1) within just a few black-box calls. Some tasks required fewer than 10 API calls to find the optimal instruction, such as for `active to passive' and `letters list', and fewer than 20 for tasks like `translation' and `diff'. It can be seen that the vast majority of tasks achieved their best reward value within the first 60 to 80 calls, demonstrating that ACING can even be used for much more constrained budgets. The choice of 165 calls was mainly based on previous work \cite{instinct, instruct}, avoiding any potential advantage that could come from optimizing this number.

\subsection{ACING with Budget Splitting}
\label{split_budget_appendix}

Due to the stochastic nature of the black-box LLM, the same instruction may yield different rewards when evaluated by the LLM. To address this, we add a mechanism for more robust decision-making. The budget $T$ is split into two parts: an exploration phase where steps 1 to 4 are repeated, and an exploitation phase where the best $p$ prompts are evaluated multiple times, $k$ times each, using the black-box LLM. The exploration phase uses $T - p \cdot k$ API calls, with the remaining calls used for exploitation. Finally, the prompt with the highest average reward across repetitions is used at test time. 
In Table \ref{tab:instruction_induction_app}, we demonstrate that ACING, with an exploration budget of $T=150$ and the remaining 15 calls allocated to uniform exploration of the top $p=5$ prompts (evaluated $k=3$ times each), achieves improved median scores across tasks and higher test accuracy on 13 tasks compared to previous work. Furthermore, it acheives a higher median score compared to ACING without splitting.

\begin{table*}[t]
    \begin{center}
    \resizebox{0.6\textheight}{!}{
    \resizebox{0.99\textwidth}{!}{
    \begin{tabular}{llcccc||cc}
    \hline
    \textbf{Category}          & \textbf{Task}             & \textbf{APE}   &   \textbf{EvoPrompt}   & \textbf{InstructZero}   & \textbf{INSTINCT} & \textbf{ACING$_{150+15}$} & \textbf{ACING$_{165}$}\\
          &          &    &    &   &  & (budget splitting) &  (main paper) \\
    \hline
    \textit{Spelling} 
     & Letters\_list             &   0.59 (0.02)   & 0.97 (0.03)    &    \textbf{1.00 (0.00)}   &    0.99 (0.01)      &   \textbf{1.00 (0.00)} &     \textbf{1.00 (0.00)}      \\
    & First\_word\_letter       &  0.00 (0.00)       &   \textbf{1.00 (0.00)}   &   \textbf{1.00 (0.00)}                            &    \textbf{1.00 (0.00)}      &     \textbf{1.00 (0.00)}      &  \textbf{1.00 (0.00) }\\
                       & Second\_word\_letter      &  0.00 (0.00) &  0.63 (0.17)  &    0.35 (0.09)                            &    0.39 (0.28)      & \textbf{0.40 (0.17)}  &  \textbf{0.70 (0.15) } \\
    \hline
    \textit{Morpho-Syntax} & Singular\_to\_plural      &  \textbf{1.00 (0.00)} &  \textbf{1.00 (0.00)}        &       0.99 (0.01)                          &    0.95 (0.03)      &     0.99 (0.01)      &  0.95 (0.03)  \\
                            & Active\_to\_passive       &   \textbf{1.00 (0.00)} &      0.99 (0.00) &      0.98 (0.01)                       &     \textbf{1.00 (0.00)}      &     \textbf{1.00 (0.00)}   & \textbf{1.00 (0.00)}     \\ 
                            & Negation                  &   0.79 (0.00)  &   \textbf{0.84 (0.02)}       &    0.65 (0.10)                            &    0.58 (0.22)      & 0.82 (0.00) &  0.71 (0.06)    \\
    \hline
    \textit{Lexical Semantics} & Antonyms                  &   0.79 (0.02)  &   0.70 (0.01)       &    0.76 (0.00)                                &   \textbf{0.84 (0.01)}       &      0.76 (0.06)  & 0.74 (0.01)                     \\
                                & Synonyms                  & 0.14 (0.01)  & 0.19 (0.07)        &    \textbf{0.22 (0.11)}                            &    0.19 (0.08)      &     0.12 (0.02) & 0.13 (0.02)       \\
                &Word\_unscrambling        &   0.54 (0.00)     &   0.44 (0.06)  &    \textbf{0.59 (0.06)}    &    0.54 (0.02)      &     \textbf{0.59 (0.05)}  & 0.50 (0.07)     \\
    \hline
    \textit{Phonetics} & Rhymes                    &  0.59 (0.01)   &  0.52 (0.05)      &    \textbf{0.99 (0.01)}    &    0.36 (0.04)      &      0.60 (0.38)  & 0.57 (0.31)      \\
    \hline
    \textit{Numerical} & Sum                       &   0.87 (0.01)    &   \textbf{1.00 (0.00)}    &   \textbf{1.00 (0.00)}     &    0.70 (0.21)      &     0.98 (0.01)  & \textbf{1.00 (0.00)}    \\
                       & Diff                      &   0.00 (0.00)   &      0.99 (0.01)     &   \textbf{1.00 (0.00)}                    &    0.93 (0.09)      &      0.97 (0.04)  &  \textbf{1.00 (0.00)}     \\
    \hline
    \textit{Knowledge} & Larger\_animal            &   0.72 (0.02)      &        0.58 (0.06)  &        0.63 (0.07)                         &    0.81 (0.09)      &   \textbf{0.86 (0.06)}  & \textbf{0.84 (0.07) }      \\
                       & Periodic\_elements        &  0.99 (0.01)      &           0.92 (0.00)   &           0.96 (0.03)                      &   \textbf{1.00 (0.00)}       &    0.98 (0.03)   & 0.98 (0.00)        \\
    \hline
    \textit{Cognitive Tasks} & Cause\_and\_effect        &   0.44 (0.09)  &   0.48 (0.10)     &    0.52 (0.09)                                &    0.55 (0.11)               & \textbf{0.76 (0.18)}  & \textbf{0.69 (0.15)}   \\
                              & Common\_concept           &   0.03 (0.02)      &    0.17 (0.00)  &    0.14 (0.04)                       &    0.09 (0.04)               &       0.10 (0.01)    & \textbf{0.19 (0.05)}    \\
                              & Object\_counting          &    0.30 (0.02) &    \textbf{0.50 (0.06)}     &    0.38 (0.06)                             &    0.40 (0.12)      & 0.48 (0.11) & 0.41 (0.03)\\
                              & Odd\_one\_out             &    0.32 (0.02)   &   \textbf{ 0.64 (0.04)}  &   0.57 (0.02)                              &    0.25 (0.18)      &     \textbf{0.59 (0.05)}  & \textbf{0.64 (0.00)}      \\
                              & Orthography\_starts\_with &    0.23 (0.01)   &    0.47 (0.02)   &    0.41 (0.09)                            & \textbf{0.54 (0.06)} &      \textbf{0.54 (0.15)}  & \textbf{0.60 (0.12) }     \\
                              & Taxonomy\_animal          &     0.02 (0.02)   &    0.38 (0.06)  &    0.67 (0.14)                       
                              &    \textbf{0.85 (0.06)}      &     0.53 (0.34)     & 0.71 (0.02) \\
& Auto\_categorization      &   \textbf{0.31 (0.01)}    &    0.20 (0.03)   &    0.29 (0.02)                                &    0.07 (0.07)               &       0.27 (0.06)  & 0.29 (0.04)      \\
&Word\_sorting             &    0.58 (0.01)   &     0.01 (0.00)   &     0.64 (0.05)                            &    0.23 (0.20)      &     \textbf{0.72 (0.02)} & \textbf{0.70 (0.03)}    \\
    \hline
    \textit{CLUE} & Sentence\_similarity      &   0.00 (0.00) &   0.05 (0.00)      &    0.10 (0.00)                             &    0.00 (0.00)      &\textbf{0.13 (0.08)} & \textbf{0.13 (0.07)}  \\
                       & Sentiment                 &  \textbf{0.90 (0.00)}   &    0.63 (0.17)    &    0.88 (0.03)                             &    0.88 (0.02)      &   0.88 (0.03)  & 0.89 (0.01)                          \\
    \hline
    \textit{Translation} 
     &Num\_to\_verbal           &   0.13 (0.02)   &   \textbf{1.00 (0.00)}    &   0.99 (0.01)                             &    \textbf{1.00 (0.00)}      &   \textbf{1.00 (0.00)}     & 0.99 (0.01)    \\
    & Translation\_en-de        &    \textbf{0.83 (0.01)}  & 0.80 (0.02)    &    0.82 (0.01)                             &    0.77 (0.02)     &      0.82 (0.01)  & 0.82 (0.01)  \\
                         & Translation\_en-es        & 0.86 (0.01)  &    0.76 (0.00)      &    0.67 (0.24)                             &   \textbf{0.89 (0.00)}     &      0.86 (0.02)   & 0.87 (0.02)   \\
                         & Translation\_en-fr        &   \textbf{0.88 (0.01)}    &   0.86 (0.00)   &   0.77 (0.06)                    
                         &   0.85 (0.02)      &    0.85 (0.02)   & 0.83 (0.01)   \\
 \hline
    \textit{Style}  & Informal\_to\_formal      &   \textbf{0.57 (0.01)}      &   0.50 (0.02)  &   0.48 (0.02)  &    0.54 (0.09)      & 0.44 (0.05) &  0.44 (0.05)\\
    \hline
    \textit{Coding} 
                       & Auto\_debugging           &   0.25 (0.00)  & 0.25 (0.00)    &    0.25 (0.00)                                &    0.07 (0.07)                &\textbf{0.29 (0.07)}   & \textbf{0.25 (0.00)}   \\

    \hline
    &median score              &       0.49        &    0.63   &    0.66                                    &     0.64            &        \textbf{0.76}       &        \textbf{0.71}          \\    
    &\# best-performing tasks  &      7       & 7       &       7                                     &       8              &        \textbf{13}     &      \textbf{13}        \\
    \hline
    \end{tabular}}}
    \end{center}
    \caption{\small Average test performance (and standard deviations) across 3 random seeds comparing \textsc{ACING} versus APE, InstructZero, EvoPrompt, and INSTINCT. The bottom rows report the median score and total number of best-performing tasks for each method.}
    \label{tab:instruction_induction_app}
\end{table*}

\subsection{ACING with Different Intrinsic (Action) Dimensions}
\label{different_dimensions}

In the main paper, we present results using actions with a dimension of $d' = 10$, following the setup of prior work. To evaluate the performance of ACING across different dimensionalities, we conducted experiments with $d' \in \{5, 10, 20, 40\}$, keeping other parameters fixed, for a budget of $165$. We report the test results over different tasks and dimensionalities for a fixed seed. The results, shown in Table \ref{tab:instruction_induction_dimension}, indicate that while the smallest dimension, $d' = 5$, recovered the best scores for some tasks, it generally has the lowest performance across most tasks. 
Furthermore, both $d' = 10$ and $d' = 20$ yield similar performance in terms of the number of best-performing tasks (9-10 tasks), indicating low sensitivity to this parameter. For the much larger dimension, $d' = 40$, the method achieved the highest number of best-performing tasks (15 tasks), demonstrating improved performance with increased dimensionality. Further increasing the dimensionality to $d' = 100$ can still yield high results, outperforming $d' \in {5, 10, 20}$. However, while it remarkably outperformed $d' = 40$ in some tasks, such as the second word letter task, synonyms, and antonyms, it only achieved 14 best-performing tasks overall, indicating similar but slightly lower performance than $d'=40$.

\begin{table*}[t]
    \begin{center}
    \resizebox{0.5\textheight}{!}{
    \resizebox{0.9\textwidth}{!}{
    \begin{tabular}{llccccc}
    \hline
    \textbf{Category}          & \textbf{Task}  & \textbf{$d^{'}=5$}        & \textbf{$d^{'}=10$}   & \textbf{$d^{'}=20$} & \textbf{$d^{'}=40$} & \textbf{$d^{'}=100$}\\
             &   &         & (main paper)   &  &  \\
    \hline
    \textit{Spelling} 
                             & Letters\_list             &      \textbf{1.00}   &      \textbf{1.00}        & 0.98  &      \textbf{1.00}  &  \textbf{1.00}  \\
                             
                            & First\_word\_letter       &   \textbf{1.00}    &    \textbf{1.00}          & 0.97  &   \textbf{1.00}   & \textbf{1.00}  \\
                            & Second\_word\_letter      &   0.23    &   0.91    &  0.30   &   0.29   &  \textbf{0.92}  \\
    \hline
    \textit{Morpho-Syntax} & Singular\_to\_plural      &  0.99    &    0.99    & \textbf{1.00}    &    \textbf{1.00} & \textbf{1.00} \\
                            & Active\_to\_passive       &   \textbf{1.00} &   \textbf{1.00}   &   \textbf{1.00}   &   \textbf{1.00} &  \textbf{1.00}  \\ 
                            & Negation                  &    0.82    &    0.80   &  \textbf{0.84}  &  0.81  & 0.70 \\
    \hline
    \textit{Lexical Semantics} & Antonyms                  &  0.73   &     0.76   &  0.76   &    0.82      & \textbf{0.84}             \\
                                & Synonyms                &      0.12   &   0.13        &     0.14   &      0.14 & \textbf{0.34}\\
                                 &Word\_unscrambling        &   0.53   &    0.54    & 0.49   &   \textbf{0.55}  &  0.43  \\
    \hline
    \textit{Phonetics} & Rhymes                    &   0.95   &   0.36    &  0.94    &      \textbf{1.00}  &  \textbf{1.00} \\
    \hline
    \textit{Numerical} & Sum                       &     0.99   &    \textbf{1.00}   &     \textbf{1.00}    &    0.99  &    \textbf{1.00} \\
                       & Diff                      &  0.89      &    \textbf{1.00}   &   \textbf{1.00}   &     \textbf{1.00} &  \textbf{1.00}   \\
    \hline
    \textit{Knowledge} & Larger\_animal            &     0.79   &     \textbf{0.94}       &  0.93       &     0.65   & 0.68   \\
                       & Periodic\_elements        &    \textbf{1.00}      &     0.98    &    0.94    &        0.98  & 0.98 \\
    \hline
    \textit{Cognitive Tasks} & Cause\_and\_effect        &   0.64   &  0.52      &   \textbf{0.92}      &   0.56  &   0.56   \\
                              & Common\_concept           &  0.12    &     \textbf{0.23}    &  0.11    &     0.12   & 0.02  \\
                              & Object\_counting          &  0.51    &     0.39    &  0.48       &  \textbf{0.59} &  0.44 \\
                              & Odd\_one\_out             &    0.60   &  0.64       & 0.64    &    \textbf{0.68}  & 0.26  \\
                              & Orthography\_starts\_with &   0.11   &     0.65     &   0.59  &     0.61 &  \textbf{0.71}   \\
                              & Taxonomy\_animal          &   0.79    &    0.68     &    0.59  &    0.85 & \textbf{0.97}  \\
                                & Auto\_categorization      &   0.30  &    0.28    &  0.13     &    \textbf{0.33} & 0.32  \\
                            &Word\_sorting             &   0.55    &     0.69       &   \textbf{0.74}    &    0.69  & 0.48  \\
    \hline
    \textit{CLUE} & Sentence\_similarity      &   0.00   &  \textbf{0.21}     &   0.00   &  0.14 & 0.07 \\
                       & Sentiment                 &   \textbf{0.91}     &  0.88    &  0.86    &  \textbf{0.91} &  0.80 \\
    \hline
    \textit{Translation} 
                         &Num\_to\_verbal           &  0.99     & \textbf{1.00}       & \textbf{1.00}  &    \textbf{1.00} & \textbf{1.00}    \\
                         & Translation\_en-de        &   \textbf{0.83}   &   0.81   &   0.81   &   0.80 &  0.81\\
                         & Translation\_en-es        &   0.89     &    0.90   &   \textbf{0.91}   &   \textbf{0.91} & 0.86  \\
                         & Translation\_en-fr        &   0.84  &   0.84   &   0.86 &    \textbf{0.88} & 0.73 \\
 \hline
    \textit{Style}  & Informal\_to\_formal      &  \textbf{0.54}   &   0.40 & 0.51   & 0.49 & 0.50 \\
    \hline
    \textit{Coding} 
                       & Auto\_debugging           &   \textbf{0.25} &   \textbf{0.25}   &  \textbf{0.25}  &   \textbf{0.25} &   \textbf{0.25} \\

    \hline
    &\# best-performing tasks  &    8     &    10    &    10     &    \textbf{15}    &  14 \\
    \hline
    \end{tabular}}}
    \end{center}
    \caption{\small Average \textsc{ACING} test performance for a fixed random seed (0) with different soft prompt dimensions $d^{'}$. The bottom row report the total number of best-performing tasks.}
    \label{tab:instruction_induction_dimension}
\end{table*}

\subsection{ACING with Different Number of Exemplars}
\label{different_exemplars}

In this section, we test ACING with a single exemplar, in contrast to the main results in the paper, which use five exemplars for ACING and all other benchmarks. For these experiments, we fix all hyperparameters as in the main paper and run tests with a budget of $165$. Intuitively, providing more exemplars to the language model should facilitate prompt learning, so five exemplars are expected to yield better prompts than a single exemplar. Our experiments, summarized in Table \ref{tab:instruction_induction_exemplars}, support this intuition. The results show that using five exemplars leads to higher test scores, as reflected in a greater number of best-performing tasks and an increase in median test scores across tasks. However, it is notable that performance did not decrease drastically with only one exemplar, suggesting that a single exemplar is sufficient to achieve decent results. In fact, across several tasks and categories (e.g., phonetics, summation, morpho-syntax, and translation), a single exemplar achieves the same performance of using five exemplars, and even outperforms the use of five exemplars in certain tasks. Nevertheless, using a single exemplar resulted in lower performance mainly in more cognitively challenging tasks, which is understandable, as more complex tasks are likely to benefit from additional exemplars.

\begin{table*}[h]
    \begin{center}
    \small
    \resizebox{0.5\textheight}{!}{
    \resizebox{0.8\textwidth}{!}{
    \begin{tabular}{llcc}
    \hline
    \textbf{Category}          & \textbf{Task}  & ACING ($|\mathcal{E}|=1$)        & ACING ($|\mathcal{E}|=5$) \\
             &  &       & (main paper) \\
    \hline
    \textit{Spelling} 
                             & Letters\_list             &  \textbf{1.00 (0.00)}     &     \textbf{1.00 (0.00)}        \\
                             
                            & First\_word\_letter       &   0.99 (0.01)   &    \textbf{1.00 (0.00)}      \\
                            & Second\_word\_letter      &  0.19 (0.09)    &      \textbf{0.70 (0.15)}  \\
    \hline
    \textit{Morpho-Syntax} & Singular\_to\_plural      &    \textbf{1.00 (0.00)}  &  0.95 (0.03)     \\
                            & Active\_to\_passive       & \textbf{1.00 (0.00)} &   \textbf{1.00 (0.00)}    \\ 
                            & Negation                  &  \textbf{0.76 (0.08)}   &  0.71 (0.06)   \\
    \hline
    \textit{Lexical Semantics} & Antonyms                  &  \textbf{0.78 (0.05)}   &   0.74 (0.01)          \\
                                & Synonyms                &  0.09 (0.03)    &  \textbf{0.13 (0.02)}  \\
                                 &Word\_unscrambling        &  0.41 (0.09)   &  \textbf{0.50 (0.07)}      \\
    \hline
    \textit{Phonetics} & Rhymes                    &    \textbf{0.89 (0.08)}  &   0.57 (0.31)   \\
    \hline
    \textit{Numerical} & Sum                       &  0.99 (0.01)   &    \textbf{1.00 (0.00)}   \\
                       & Diff                      &  0.99 (0.01)    &     \textbf{1.00 (0.00)}      \\
    \hline
    \textit{Knowledge} & Larger\_animal            &  0.63 (0.17)     &     \textbf{0.84 (0.07)}       \\
                       & Periodic\_elements        &  0.91 (0.08)    & \textbf{0.98 (0.00)}\\
    \hline
    \textit{Cognitive Tasks} & Cause\_and\_effect        &  0.51 (0.08) &  \textbf{0.69 (0.15)}      \\
                              & Common\_concept           &  0.16 (0.11)  &   \textbf{0.19 (0.05)}       \\
                              & Object\_counting          &  0.26 (0.06)   &   \textbf{0.41 (0.03)}       \\
                              & Odd\_one\_out             &   \textbf{0.64 (0.02)}  &  \textbf{0.64 (0.00)} \\
                              & Orthography\_starts\_with &  0.06 (0.05)   &   \textbf{0.60 (0.12)}        \\
                              & Taxonomy\_animal          &  0.63 (0.06)   &  \textbf{0.71 (0.02)} \\
                                & Auto\_categorization      &    0.01 (0.01)  &  \textbf{0.29 (0.04)}    \\
                            &Word\_sorting             &    \textbf{0.70 (0.02)}   &     \textbf{0.70 (0.03)}         \\
    \hline
    \textit{CLUE} & Sentence\_similarity      &   0.07 (0.05)   &      \textbf{0.13 (0.07)}  \\
                       & Sentiment                 &     0.70 (0.12)   &   \textbf{0.89 (0.01)}     \\
    \hline
    \textit{Translation} 
                         &Num\_to\_verbal           &   \textbf{1.00 (0.00)}   &  0.99 (0.01)       \\
                         & Translation\_en-de        &  0.72 (0.11)    &    \textbf{0.82 (0.01)}    \\
                         & Translation\_en-es        &   \textbf{0.88 (0.00)}     & 0.87 (0.02)\\
                         & Translation\_en-fr        &  0.16 (0.04)  &  \textbf{0.83 (0.01)}   \\
 \hline
    \textit{Style}  & Informal\_to\_formal      &  0.42 (0.05) & \textbf{0.44 (0.05)}  \\
    \hline
    \textit{Coding} 
                       & Auto\_debugging           &  \textbf{0.25 (0.00)}  & \textbf{0.25 (0.00)}    \\

    \hline
    &median score              &  0.67     &   \textbf{0.71}      \\    
    &\# best-performing tasks  &    11     &   \textbf{24}      \\
    \hline
    \end{tabular}}}
    \end{center}
    \caption{\small Average \textsc{ACING} test performance (and standard deviations) across 3 random seeds comparing 1 exemplar versus 5 exemplars. The bottom rows report the median score and total number of best-performing tasks.}
    \label{tab:instruction_induction_exemplars}
\end{table*}

\subsection{ACING with Different White-box models}
\label{different_white_llms}

In this section, we evaluate the impact of the choice of white-box model on the ACING method. Specifically, we applied ACING for instruction learning with a GPT-3.5-turbo as the black-box LLM (as in the main paper), but using different white-box models. In the main paper, we reported ACING with Vicuna-$13\text{B-v}1.3$; in Table \ref{tab:instruction_induction_white_box_diff}, we further test it with WizardLM-$13\text{B-v}1.2$.  As shown in the table, changing the white-box model results in slight variations in performance. WizardLM achieved a higher median test score across all tasks and excelled in a greater number of top-performing tasks.

\begin{table*}[t]
    \begin{center}
    \resizebox{0.5\textheight}{!}{
    \resizebox{0.99\textwidth}{!}{
    \begin{tabular}{llcc}
    \hline
    \textbf{Category}          & \textbf{Task}             & \textbf{ACING (Vicuna)} & \textbf{ACING (WizardLM)} \\
        &            & (main paper) &  \\
    \hline
    \textit{Spelling} 
     & Letters\_list                &     \textbf{1.00 (0.00)}       &    \textbf{1.00 (0.00)}   \\
    & First\_word\_letter           &     \textbf{1.00 (0.00)}   &    \textbf{1.00 (0.00)}      \\
    & Second\_word\_letter                &     \textbf{0.70 (0.15)}   &    0.36 (0.18)     \\
    \hline
    \textit{Morpho-Syntax} & Singular\_to\_plural       &     0.95 (0.03)   &     \textbf{0.99 (0.00)}       \\
                            & Active\_to\_passive       &     \textbf{1.00 (0.00)} &     \textbf{1.00 (0.00)}  \\ 
                            & Negation                     & 0.71 (0.06)   &    \textbf{0.83 (0.00)}  \\
    \hline
    \textit{Lexical Semantics} & Antonyms       &      0.74 (0.01)      &  \textbf{ 0.81 (0.02)}                    \\
                                & Synonyms    &     \textbf{0.13 (0.02)}     &    0.12 (0.03)       \\
                &Word\_unscrambling   &     0.50 (0.07)     &    \textbf{0.57 (0.05)}        \\
    \hline
    \textit{Phonetics} & Rhymes    &      0.57 (0.31)   &    \textbf{0.97 (0.04)}         \\
    \hline
    \textit{Numerical} & Sum    &     \textbf{1.00 (0.00)}     &    \textbf{1.00 (0.00)}      \\
                       & Diff    &      \textbf{1.00 (0.00)}     &    \textbf{1.00 (0.00) }      \\
    \hline
    \textit{Knowledge} & Larger\_animal       &  0.84 (0.07)     &    \textbf{0.94 (0.01)}      \\
                       & Periodic\_elements       &    \textbf{0.98 (0.00)}    &   0.97 (0.02)         \\
    \hline
    \textit{Cognitive Tasks} & Cause\_and\_effect     & 0.69 (0.15)  &    \textbf{0.76 (0.20)}   \\
                              & Common\_concept &       0.19 (0.05)      &    \textbf{0.21 (0.05)}  \\
                              & Object\_counting & 0.41 (0.03)   &   \textbf{0.46 (0.07)} \\
                              & Odd\_one\_out          &     \textbf{0.64 (0.00)}  &    0.56 (0.11)       \\
                              & Orthography\_starts\_with  & 0.60 (0.12)   & \textbf{0.62 (0.03) }    \\
                              & Taxonomy\_animal   &     \textbf{0.71 (0.02)}    &    0.60 (0.32)   \\
& Auto\_categorization                &       0.29 (0.04)    &    \textbf{0.35 (0.03)}      \\
&Word\_sorting                     &     \textbf{0.70 (0.03)}   &    0.61 (0.02)    \\
    \hline
    \textit{CLUE} & Sentence\_similarity     & 0.13 (0.07)  &  \textbf{ 0.22 (0.04)}    \\
                       & Sentiment                     &   0.89 (0.01)    &    \textbf{0.90 (0.02)  }                        \\
    \hline
    \textit{Translation} 
     &Num\_to\_verbal      &   0.99 (0.01)      &    \textbf{1.00 (0.00)}      \\
    & Translation\_en-de       &      \textbf{0.82 (0.01) }&    0.81 (0.01)    \\
                         & Translation\_en-es     &     \textbf{0.87 (0.02)}   &   0.61 (0.38)     \\
                         & Translation\_en-fr                       
                             &   \textbf{0.83 (0.01)}  &   \textbf{0.83 (0.05)}      \\
 \hline
    \textit{Style}  & Informal\_to\_formal       & \textbf{0.44 (0.05)} &    0.32 (0.19) \\
    \hline
    \textit{Coding} 
                       & Auto\_debugging & 0.25 (0.00)     &   \textbf{ 0.38 (0.10)}  \\

    \hline
    &median score            &        0.71     &     \textbf{0.79}                        \\    
    &\# best-performing tasks              &      15          &         \textbf{21}                  \\
    \hline
    \end{tabular}}}
    \end{center}
    \caption{\small Average \textsc{ACING} test performance (and standard deviations) across 3 random seeds using Vicuna and WizardLM as white-box models. The bottom rows report the median score and total number of best-performing tasks.}
    \label{tab:instruction_induction_white_box_diff}
\end{table*}

\section{Demonstrations with Human Instructions}
\label{sec:best_instruction}

To contextualize the performance of \textsc{ACING}, we compare its best-learned instructions against human-written instructions curated by \citet{honovich-etal-2023-instruction}. Table~\ref{tab:human_performance_appendix} presents a representative subset of tasks, categorized by linguistic and semantic attributes, along with input–output demonstrations, human instructions, and corresponding performance scores. While human instructions often perform strongly, \textsc{ACING} matches or exceeds them in the majority of cases, particularly on tasks like Antonyms, Rhymes, and Sentence Similarity, where learned instructions yield notable improvements. The comparison underscores \textsc{ACING}’s capacity not only to automate instruction crafting but also to outperform carefully designed human-written prompts across diverse task types. Summary statistics at the bottom of the table show that \textsc{ACING} achieves a higher average and median score, and wins on a greater number of tasks overall.

\begin{table*}[t]
\small
\centering
\resizebox{0.62\textheight}{!}{
\begin{tabular}{@{}p{0.1\textwidth}@{}p{0.18\textwidth}@{}p{0.19\textwidth}p{0.35\textwidth}p{0.06\textwidth}p{0.06\textwidth}@{}}
\toprule
\textbf{Category}                           & \textbf{Task}                                                        & \textbf{Demonstration} & \textbf{Human Instruction \cite{honovich-etal-2023-instruction}} & \textbf{Human Score} & \textbf{Our Score} \\
\midrule
\textit{Spelling} & First\_word\_letter & cat $\rightarrow$  c  & Extract the first letter of the input word. & \textbf{1.00 (0.00)} & \textbf{1.00 (0.00)}\\
\cmidrule{2-6}
 & Second\_word\_letter & cat $\rightarrow$  a & Extract the second letter of the input word. & \textbf{0.96 (0.00)} & 0.92 (0.00)\\
\cmidrule{2-6}
 & Letters\_list  & cat $\rightarrow$  c a t & Break the input word into letters, separated by spaces. & \textbf{1.00 (0.00)} & \textbf{1.00 (0.00)}\\
\midrule
\textit{Morpho-}

\textit{syntax} 
& Singular\_to\_plural                   & cat $\rightarrow$  cats  & Convert the input word to its plural form. & \textbf{1.00 (0.00)} & \textbf{1.00 (0.00)}\\
\cmidrule{2-6} 
                                   & Active\_to\_passive              &
The artist introduced the scientist. $\rightarrow$  The scientist was introduced by the artist. & Write the input sentence in passive form.&  \textbf{1.00 (0.00)} & \textbf{1.00 (0.00)}\\
\midrule
\textit{Syntax}                             & Negation     & Time is finite $\rightarrow$  Time is not finite.  & Negate the input sentence. & 0.81 (0.00) & \textbf{0.82 (0.00)} \\
\midrule
\textit{Lexical} 

\textit{Semantics} & Antonyms    & won $\rightarrow$ lost   & Write a word that means the opposite of the input word. & 0.70 (0.00) & \textbf{0.83 (0.00)}\\
\cmidrule{2-6}
   & Synonyms    & alleged $\rightarrow$  supposed   & Write a word with a similar meaning to the input word.&  \textbf{0.14 (0.01)} & 0.13 (0.00)\\
\midrule
\textit{Phonetics}                          & Rhymes       & sing $\rightarrow$  ring   & Write a word that rhymes with the input word. &   0.61 (0.01)   & \textbf{1.00 (0.00)}    \\
\midrule
\textit{Knowledge}                    & Larger\_animal & koala, snail $\rightarrow$  koala & Write the larger of the two given animals.   &    \textbf{0.94 (0.00)}  & \textbf{0.94 (0.00)}    \\
\midrule
\textit{Semantics} 
& Cause\_and\_effect  & Sentence 1: The soda went flat. Sentence 2: The bottle was left open. $\rightarrow$  The bottle was left open.& Find which of the two given cause and effect sentences is the cause. & \textbf{0.97 (0.02)} & 0.90 (0.02) \\
\cmidrule{2-6}
& Common\_concept & guitars, pendulums, neutrinos $\rightarrow$  involve oscillations. & Find a common characteristic for the given objects. & \textbf{0.11 (0.01)} & \textbf{0.11 (0.00)} \\
\midrule
\textit{Style} & Informal\_to\_formal & Please call once you get there $\rightarrow$  Please call upon your arrival. & Rephrase the sentence in formal language. & \textbf{0.63 (0.00)} & 0.50 (0.00)\\
\midrule
\textit{Numerical}         & Sum     & 22 10 $\rightarrow$  32       & Sum the two given numbers.  & \textbf{1.00 (0.00)} & \textbf{1.00 (0.00)}\\
\cmidrule{2-6}
   & Diff  & 32 22 $\rightarrow$  10   & Subtract the second number from the first. &  \textbf{1.00 (0.00)} & \textbf{1.00 (0.00)}\\
\cmidrule{2-6}
              & Num\_to\_Verbal & 26 $\rightarrow$  twenty-six & Write the number in English words. & \textbf{1.00 (0.00)} & \textbf{1.00 (0.00)}\\
\midrule
\textit{Multi-}

\textit{lingual} & Translation\_en-de   & game $\rightarrow$  Spiel & Translate the word into German.  &  0.81 (0.00)& \textbf{0.84 (0.00)}\\

\cmidrule{2-6}
& Translation\_en-es    & game $\rightarrow$  juego & Translate the word into Spanish. &  \textbf{0.89 (0.00)} & 0.88 (0.00)\\

\cmidrule{2-6}
& Translation\_en-fr   & game $\rightarrow$  jeu  & Translate the word into French.  & 0.86 (0.00) & \textbf{0.87 (0.00)}\\

\midrule
\textit{GLUE} & Sentiment & The film is small in scope, yet perfectly formed. $\rightarrow$  positive & Determine whether a movie review is positive or negative. & 0.89 (0.01) & \textbf{0.91 (0.00)} \\
\cmidrule{2-6}
& Sentence\_similarity & Sentence 1: A man is smoking. Sentence 2: A man is skating. $\rightarrow$  0 - definitely not & Rate the semantic similarity of two input sentences on a scale of 0 - definitely not to 5 - perfectly. & 0.00 (0.00) & \textbf{0.21 (0.00)} \\
\bottomrule 
 & &  & \textit{median score} & 0.89 & \textbf{0.91} \\
  & &  & \textit{average score} & 0.78 & \textbf{0.80} \\
 & &  & \textit{\# best-performing tasks} & 14 & \textbf{16} \\
 \bottomrule \\
\end{tabular}}
\caption{Classified tasks into categories from the instruction-induction datasets. For each task, we provide a corresponding demonstration, with $\rightarrow$ separating the input from the output, along with its respective human instruction as proposed in \cite{honovich-etal-2023-instruction}. We tested these instructions, report their test scores (mean over 3 runs, standard deviation in parentheses), and compare them to our best test scores using \textsc{ACING} with Vicuna-13B as the white-box model.}
\label{tab:human_performance_appendix}
\end{table*}

\section{Assessing Instruction Clarity: Human and Readability Analyses}
\label{sec:instruction-clarity}

To assess the interpretability and accessibility of the instructions generated by \textsc{ACING}, we conduct both a human evaluation and an automated readability analysis.

\subsection{Human Evaluation}

We conducted a human annotation study across all instruction-induction tasks reported in \cref{tab:human_performance_main}.

A total of 26 participants (none of whom are paper co-authors) volunteered to independently rate the clarity, coherence, and task faithfulness of instructions produced by \textsc{ACING} using a 5-point Likert scale. The following question was used to guide ratings:

\begin{quote}
\textit{Does the instruction clearly describe what is happening in the demonstration? Could a language model complete the task correctly by following this instruction alone?}
\end{quote}

Each instruction was presented alongside:
\begin{itemize}
\item The task name and category
\item An input–output demonstration pair
\item The corresponding \textsc{ACING}-generated instruction
\end{itemize}

\paragraph{Results.} The results of the human evaluation are summarized below:
\begin{itemize}
\item \textbf{Maximum score:} 4.92
\item \textbf{Third quartile (Q3):} 4.50
\item \textbf{Median (Q2):} 3.90
\item \textbf{First quartile (Q1):} 2.83
\item \textbf{Mean score:} 3.66 \ (\textit{SD} = 1.02)
\end{itemize}

These results indicate that the majority of instructions are perceived as clear and well-aligned with the task by human annotators.

\subsection{Automated Readability Analysis}

To complement the human study, we applied three established readability formulas to quantify the linguistic accessibility of the 30 generated instructions in \cref{tbl:best_prompt_instruction_induction}:

\begin{itemize}
\item \textbf{Flesch Reading Ease (FRE)} \cite{flesch1948new}: Scores range from 0 to 100, with higher scores indicating greater ease of reading.
\item \textbf{Flesch-Kincaid Grade Level (FKG)} \cite{kincaid1975derivation}: Maps readability to U.S. school grade levels, with lower scores indicating simpler text.
\item \textbf{Coleman-Liau Index (CLI)} \cite{coleman1975computer}: A character-based grade-level readability metric.
\end{itemize}

\paragraph{Findings.} Readability scores across instructions show the following trends:
\begin{itemize}
\item Most instructions fall within the \textbf{FRE range of 60–95}, with a median 70.8, indicating that they are accessible to a general audience.
\item \textbf{FKG scores} mostly range between 3 and 9, with a median 7.0, consistent with middle school to early high school reading levels.
\item \textbf{CLI scores} mostly range between 2–10, with a median of 7.3. This aligns with the FKG analysis, indicating that instructions are generally suitable for readers with middle school to early high school reading levels.
\end{itemize}

Overall, these results confirm that \textsc{ACING} generates instructions that are not only effective—according to human judgment—but also accessible, as measured by standardized readability metrics. The full set of readability scores is included in \cref{tab:readability_scores_full}.

\begin{table*}[t]
\small
\centering
\begin{tabular}{p{11.5cm}ccc}
\toprule
\textbf{Instruction} & \textbf{FRE} $\uparrow$ & \textbf{FKG} $\downarrow$ & \textbf{CLI} $\downarrow$ \\
\midrule
Change the input to match the output, but the output is already in the passive voice & 74.3 & 6.9 & 7.00 \\
Take a word and change it to its opposite & 94.3 & 2.3 & 2.18 \\
Match the input to the output, and the answer is:\$Input\$: Nature Nanotechnology, Annual Review of Biochemistry, and The Lancet Neurology  \$Output\$: top journals \$Input\$: Jeans, Tops, and Suits  \$Output\$: Apparel & 15.6 & 18.5 & 14.53 \\
Input the code into a Python interpreter and observe the output & 49.5 & 9.1 & 9.45 \\
Find the sentence that is the cause of the effect in the pair of sentences & 84.5 & 5.2 & 5.43 \\
Make a connection between the input and output, but the connection is not clear & 65.7 & 7.6 & 9.01 \\
Find the difference between the two numbers & 66.8 & 5.7 & 10.63 \\
Create a function that takes a string as input and returns the first letter of the first word in the string & 80.8 & 7.2 & 6.82 \\
Convert the input into output using the same word order and with the same meaning & 67.5 & 7.6 & 8.13 \\
Create a program that takes two animals as input and outputs the animal that is bigger & 58.4 & 9.1 & 8.09 \\
Input the word ``year'' and the output was ``y e a r'' & 96.0 & 2.9 & -1.35 \\
Flip the truth value of the statements in the input & 86.7 & 3.7 & 5.60 \\
Convert numbers to words & 75.9 & 3.7 & 7.25 \\
Provide a number that represents how many items are in the input & 60.7 & 7.8 & 7.35 \\
Find the word that does not belong in each group based on the given words & 95.7 & 3.6 & 5.04 \\
Find a word in the text that starts with the letter provided and to output that word & 85.1 & 5.6 & 5.66 \\
Find the name of the element based on its atomic number & 72.6 & 5.9 & 5.24 \\
Input the word that the program thought I was inputting and then output the word that program thought I was inputting & 76.7 & 7.8 & 9.58 \\
Input a word and output the letter that corresponds to the second letter in that word & 69.0 & 7.6 & 7.72 \\
Find a sentence pair that is probably not similar, and the output is 3 - probably & 61.9 & 8.4 & 6.97 \\
Classify each input as positive or negative based on the assessment of the corresponding movie & 33.7 & 12.3 & 13.16 \\
Add the suffix -s to the end of the word to make it plural & 95.9 & 3.4 & 0.31 \\
Find the sum of the two numbers & 103.0 & 0.6 & 0.69 \\
Input a word that is a synonym for the word that was output & 83.0 & 4.9 & 2.89 \\
Make the AI generate a sequence of animals based on the input provided & 57.0 & 8.5 & 7.80 \\
Provide a translation for each word in the English text into German & 67.8 & 6.8 & 8.80 \\
Translate the words from English to Spanish, but I noticed that some of the translations are not accurate & 66.4 & 8.5 & 10.59 \\
Create a program that would take an English word as input and output its French equivalent & 63.7 & 8.4 & 9.54 \\
Output the words in alphabetical order, but the output is not in alphabetical order & 35.5 & 11.8 & 10.67 \\
Convert the input to a word that is a common English word & 81.9 & 4.8 & 3.97 \\
\midrule
\textit{Median} & 70.8 & 7.0 & 7.3 \\
\bottomrule
\end{tabular}
\caption{Readability analysis of the best instructions generated by \textsc{ACING}, measured using Flesch Reading Ease (FRE) \cite{flesch1948new}, Flesch-Kincaid Grade Level (FKG) \cite{kincaid1975derivation}, and Coleman-Liau Index (CLI) \cite{coleman1975computer}. Higher FRE and lower FKG/CLI indicate easier-to-understand instructions.}
\label{tab:readability_scores_full}
\end{table*}

\section{Our Best Learned Instructions}
\label{our_best_instructions_appendix}

In this section, we present the best-learned instructions discovered by \textsc{ACING} for each of the 30 instruction induction tasks. These instructions were generated using Vicuna-13B as the white-box model and optimized to maximize performance on the corresponding black-box evaluation. Table~\ref{tbl:best_prompt_instruction_induction} showcases the resulting instructions alongside their test scores, which reflect the accuracy or task-specific metric obtained on held-out examples. The diversity and clarity of the instructions demonstrate \textsc{ACING}’s ability to synthesize task-relevant, semantically grounded prompts that elicit strong responses from black-box LLMs. Notably, several tasks achieve perfect scores, while others expose task-specific challenges (e.g., common\_concept and synonyms), highlighting the varying difficulty across the instruction spectrum.

\begin{table*}[t]
\small
\begin{center}
\resizebox{0.64\textheight}{!}{%
\begin{tabular}{lp{9cm}c}
\hline
\textbf{Task} & \textbf{Best instruction} & Test Score \\ 
\hline
active\_to\_passive& Change the input to match the output, but the output is already in the passive voice & 1.00 \\ \hline
 antonyms& Take a word and change it to its opposite & 0.82 \\ \hline
 auto\_categorization& Match the input to the output, and the answer is:$Input$: Nature Nanotechnology, Annual Review of Biochemistry, and The Lancet Neurology  $Output$: top journals $Input$: Jeans, Tops, and Suits  $Output$: Apparel & 0.34\\ \hline
 auto\_debugging& Input the code into a Python interpreter and observe the output & 0.375 \\ \hline
 cause\_and\_effect& Find the sentence that is the cause of the effect in the pair of sentences & 0.92 \\ \hline
 common\_concept& Make a connection between the input and output, but the connection is not clear & 0.11 \\ \hline
 diff& Find the difference between the two numbers & 1.00 \\ \hline
 first\_word\_letter& Create a function that takes a string as input and returns the first letter of the first word in the string & 1.00\\ \hline
 informal\_to\_formal& Convert the input into output using the same word order and with the same meaning & 0.50 \\ \hline
 larger\_animal&  Create a program that takes two animals as input and outputs the animal that is bigger & 0.94 \\ \hline
 letters\_list& Input the word "year" and the output was "y e a r" & 1.00 \\ \hline
 negation& Flip the truth value of the statements in the input & 0.82 \\ \hline
 num\_to\_verbal&  Convert numbers to words & 1.00 \\ \hline
 object\_counting& Provide a number that represents how many items are in the input & 0.55  \\ \hline
 odd\_one\_out& Find the word that does not belong in each group based on the given words & 0.64 \\ \hline
 orthography\_starts\_with& Find a word in the text that starts with the letter provided and to output that word & 0.71 \\ \hline
 periodic\_elements& Find the name of the element based on its atomic number & 1.00 \\ \hline
 rhymes& Input the word that the program thought I was inputting and then output the word that program thought I was inputting & 1.00\\ \hline
 second\_word\_letter& Input a word and output the letter that corresponds to the second letter in that word & 0.91 \\ \hline
 sentence\_similarity&  Find a sentence pair that is probably not similar, and the output is 3 - probably & 0.21 \\ \hline
 sentiment& Classify each input as positive or negative based on the assessment of the corresponding movie & 0.90 \\ \hline
 singular\_to\_plural&  Add the suffix -s to the end of the word to make it plural & 1.00 \\ \hline
 sum& Find the sum of the two numbers & 1.00\\ \hline
 synonyms& Input a word that is a synonym for the word that was output & 0.13 \\ \hline
 taxonomy\_animal& Make the AI generate a sequence of animals based on the input provided & 0.75 \\ \hline
 translation\_en-de&  Provide a translation for each word in the English text into German & 0.84 \\ \hline
 translation\_en-es& Translate the words from English to Spanish, but I noticed that some of the translations are not accurate & 0.88 \\ \hline
 translation\_en-fr& Create a program that would take an English word as input and output its French equivalent & 0.87 \\ \hline
 word\_sorting& Output the words in alphabetical order, but the output is not in alphabetical order & 0.73 \\ \hline
 word\_unscrambling&  Convert the input to a word that is a common English word & 0.63\\
\hline
\end{tabular}
}
\end{center}
\caption{
The best instruction discovered by \textsc{ACING} for all the 30 instruction-induction tasks using with Vicuna-13B as the white-box model.}
\label{tbl:best_prompt_instruction_induction}
\end{table*}

\section{Use of AI Assistance.} 

We used AI assistants (e.g., ChatGPT) in a limited and supporting role during the preparation of this paper. Specifically, we used AI tools to assist with editing text for clarity and code debugging. All core ideas, algorithms, experiments, results, analyses, and technical writing were fully developed and executed by the authors. No AI system contributed to scientific decisions, modeling choices, or interpretation of findings.

\end{document}